\documentclass{isprs} 
\usepackage[table]{xcolor}
\usepackage{setspace}
\usepackage{geometry} 
\usepackage{epstopdf}
\usepackage[labelsep=period]{caption}  
\usepackage[british]{babel} 
\usepackage[hang]{footmisc}
\usepackage{natbib}
\usepackage{subcaption}
\usepackage{siunitx}
\usepackage{graphicx}
\usepackage{arydshln} 
\usepackage{booktabs}
\usepackage{multirow}
\usepackage[export]{adjustbox}
\usepackage{enumitem}
\usepackage{svg}
\usepackage[hang]{footmisc}
\usepackage{url}
\usepackage{amssymb}
\usepackage{amsmath}
\usepackage{csquotes}
\usepackage{array}

\usepackage{threeparttable}
\usepackage{makecell}
\usepackage{xspace}

\usepackage{todonotes}

\newcommand{\tablesqueeze}{\renewcommand{\arraystretch}{0.95}\setlength{\tabcolsep}{4pt}}

\definecolor{ac}{HTML}{00C800}
\definecolor{au}{HTML}{4169E1}
\definecolor{ic}{HTML}{DC0000}
\definecolor{iu}{HTML}{FF8C00}

\begin{document}

\title{Uncertainty Quality of VGGT: An Analysis on the DTU Benchmark Dataset}
\date{}


\author{\begin{tabular}{c}
         Markus Hillemann\textsuperscript{1}, Robert Langend\"orfer\textsuperscript{1}, Steven Landgraf\textsuperscript{1}, Markus Ulrich\textsuperscript{1}
     \end{tabular}
 }

\address{\textsuperscript{1}Institute of Photogrammetry and Remote Sensing, Karlsruhe Institute of Technology, Germany - \\(markus.hillemann@, steven.landgraf@, robert.langendoerfer@, markus.ulrich@)kit.edu
}



\abstract{Visual Geometry Grounded Transformer (VGGT) has already attracted a great deal of attention in a short period of time, not least due to the Best Paper Award at CVPR-2025. Similar to DUSt3R and MASt3R, VGGT aims to bring about a paradigm shift by replacing established methods like bundle adjustment and feature matching with a simple, unified, feed-forward neural network that predicts camera poses, depth maps, and dense 3D structure directly from multiple images of a scene in a few seconds. A key aspect is its ability to process an arbitrary number of views consistently in a single forward pass without any post-processing or iterative optimization. For photogrammetry, this opens new possibilities for real-time, scalable, and accessible 3D reconstruction. In this context, not only high reconstruction accuracy but also high-quality uncertainty estimates are crucial, as they foster trust and enable robust quality assurance. This paper therefore investigates the quality of VGGT’s uncertainty predictions. The analysis identifies an effective confidence threshold for filtering VGGT’s raw output and demonstrates that enhancing uncertainty quality holds strong potential for improving the accuracy of its 3D reconstructions.}

\keywords{3D Reconstruction, 3D Foundation Models, Feed Forward, Multi-View Stereo, Uncertainty Estimation}

\maketitle


\section{Introduction}\label{sec:Introduction}

\begin{figure}
    \centering
    \begin{subfigure}[t]{0.28\linewidth}
        \includegraphics[width=\textwidth, trim=11cm 8cm 8cm 0cm, clip]{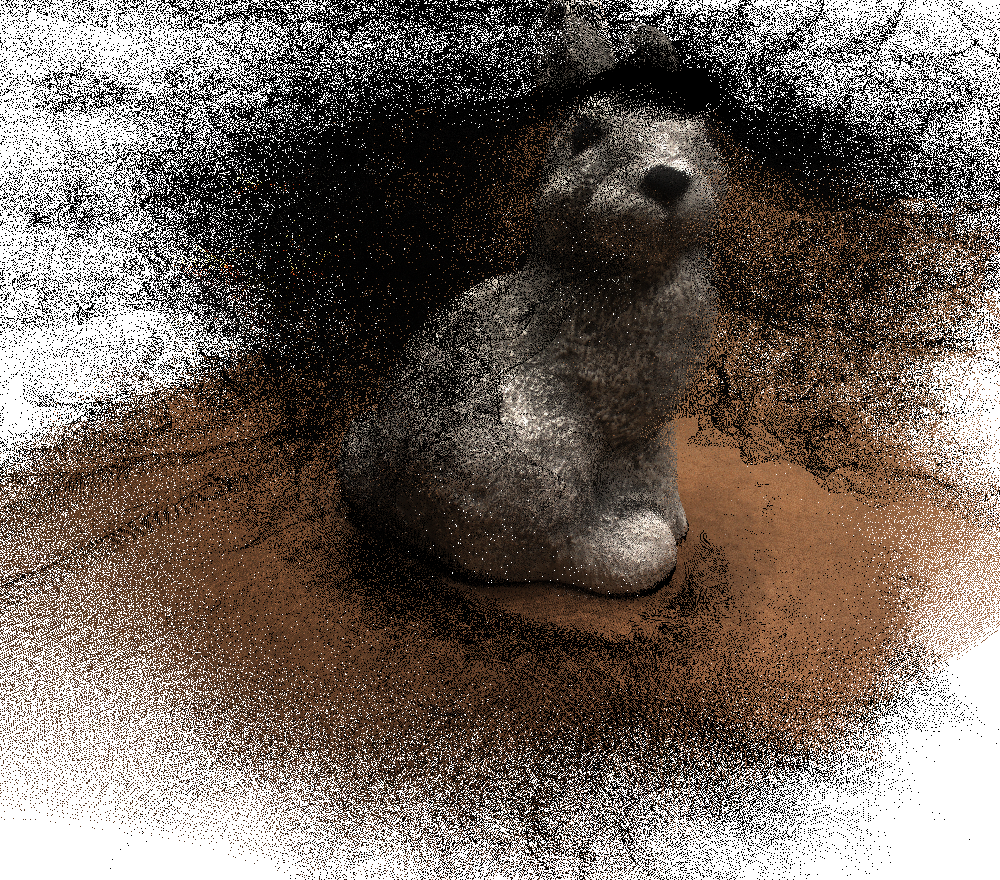}
    \end{subfigure}
    \hspace{0.5cm}
    \begin{subfigure}[t]{0.28\linewidth}
        \includegraphics[width=\textwidth, trim=11cm 8cm 8cm 0cm, clip]{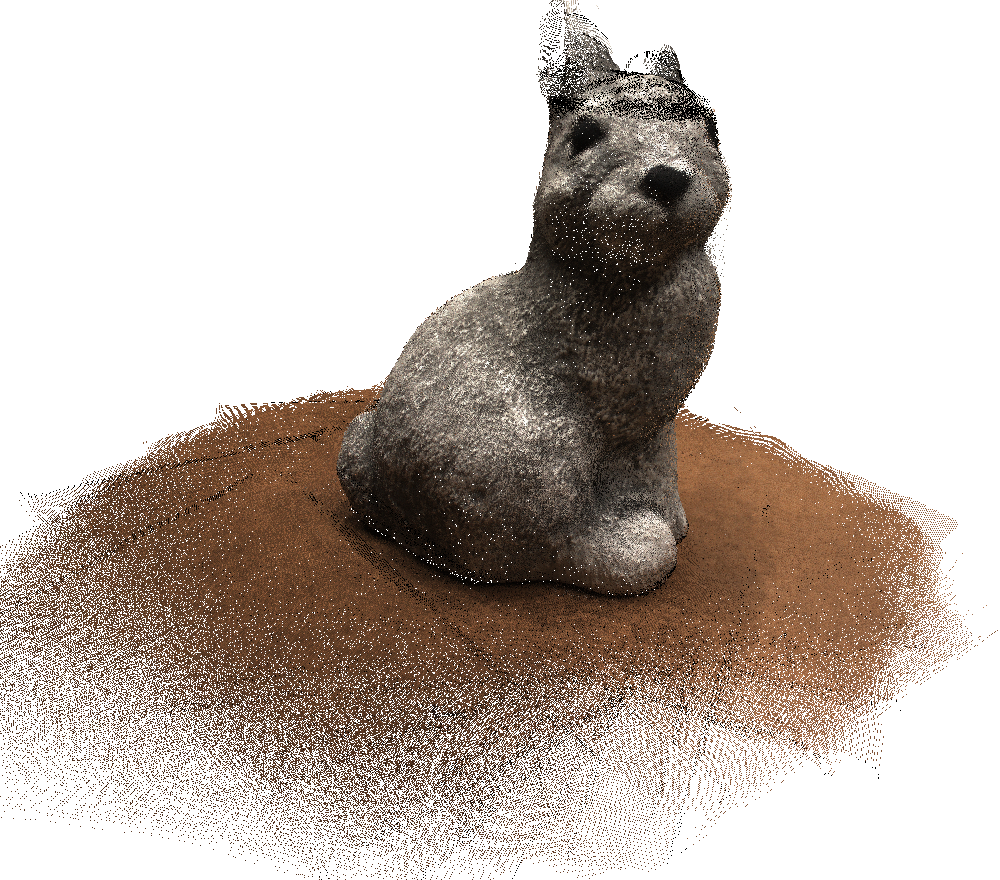}
    \end{subfigure}    
    \caption{Qualitative effect of filtering the raw output of VGGT with a confidence threshold. Left: raw VGGT point maps, right: filtered with a confidence threshold of 2.0.}
    \label{fig:conf_filter}
\end{figure}

Visual Geometry Grounded Transformer (VGGT) \citep{wang2025vggt} has received substantial attention for pushing feed-forward, learning-based 3D reconstruction towards a unified paradigm. Similar to DUSt3R \citep{wang2024dust3r} and MASt3R \citep{leroy2024grounding}, VGGT aims to replace classical, multi-stage SfM+MVS pipelines with a single forward pass that jointly estimates camera poses, depth maps, and dense 3D structure from multiple images. A key innovation is its ability to process an arbitrary number of views consistently without explicit post-optimization, which is particularly appealing for photogrammetric use cases requiring scalability and fast processing. In this context, not only geometric accuracy but also reliable uncertainty estimates are essential for trustworthy operation and rigorous quality assurance \citep{landgraf2025rethinking}. Moreover, uncertainty predictions can be used to improve 3D reconstruction accuracy, e.g., simply by filtering the redundant output of VGGT as depicted in Fig.\@~\ref{fig:conf_filter}.

Despite VGGT’s impressive progress in unifying multi-view reconstruction \citep{wang2025vggt}, the reliability of its uncertainty estimates remains largely unexplored. As noted by \citet{zhang2025review}, “another critical and underexplored area is uncertainty quantification. For safety-critical applications such as robotics, principled methods for estimating the reliability of the reconstructed geometry are essential”. Existing works have primarily focused on improving geometric accuracy and generalization, while overlooking the potential of leveraging model-intrinsic uncertainties as practical indicators of reliability or failure. In particular, it remains unclear whether VGGT’s uncertainty already encodes meaningful information about reconstruction quality. This motivates our study, which systematically investigates how useful VGGT’s native uncertainty estimates are out of the box without introducing additional uncertainty quantification mechanisms.

We deliberately evaluate the feed-forward variant of VGGT, i.e., the raw predictions without post-hoc bundle adjustment, because only then are the reported uncertainties directly interpretable and not altered by subsequent optimization. 

Our contributions are threefold:
\vspace{-0.8\baselineskip}
\begin{itemize}[nosep]
    \item We present the first systematic analysis of VGGT’s uncertainty quality.  We perform the evaluation on the DTU benchmark and consider both reconstruction branches, via point maps and via depth maps.
    \item We show that a causally meaningful confidence threshold of 2.0 effectively filters the raw outputs and improves the accuracy–completeness trade-off across scenes.
    \item We provide a comprehensive evaluation of the uncertainties with established metrics like PAvPU \citep{mukhoti2018evaluating} and AUSE \citep{ilg2018uncertainty}, revealing potential for improvement. Enhancing uncertainty quality can also contribute to improved reconstruction accuracy.
\end{itemize}

In summary, we investigate whether VGGT’s native aleatoric uncertainties already encode actionable signals for photogrammetric quality assessment, and how such signals can be leveraged in practice. The remainder of the paper reviews related work (Section~\ref{sec:RW}), introduces the basics of VGGT and its uncertainty formulation (Section~\ref{sec:basics}), details the evaluation setup (Section~\ref{sec:setup}), and presents results on the DTU dataset \citep{aanaes2016large}, including the quality of VGGTs 3D reconstruction as well as its uncertainty quality (Section~\ref{sec:Experiments}).

\section{Related work}\label{sec:RW}
\textbf{Traditional Structure from Motion and Multi-view Stereo.} Reconstructing 3D scenes from a set of 2D images has been one of the most fundamental tasks in photogrammetry \citep{zhou2024comprehensive}. Thereby, Structure from Motion (SfM) recovers the camera orientation, i.e., interior and exterior orientation, and a sparse 3D point cloud by detecting and matching local features across multiple views and solving a bundle-adjustment optimization \citep{hartley2003multiple,schonberger2016structure}. Multi-view Stereo (MVS) takes this process one step further by densely reconstructing the geometry of a scene from multiple overlapping images \citep{seitz2006comparison,stathopoulou2023survey}. Even though traditional SfM + MVS pipelines, e.g., COLMAP, \citep{schonberger2016structure,schonberger2016pixelwise}, can achieve a very high accuracy and completeness, they typically require images with generous overlap as well as clearly textured areas, and require lengthy processing times \citep{liu2025survey,wu2025evaluation}.

\textbf{Feed-forward Learning-based 3D Reconstruction.} Recent learning-based approaches have begun to replace the traditional SfM + MVS pipelines with end-to-end networks. While earlier work used learning for individual parts, for example, SuperPoint \citep{detone2018superpoint} and SuperGlue \citep{sarlin2020superglue} for keypoint detection and feature matching, they still relied on the iterative and often fragile nature of the SfM stage \citep{zhang2025review}. However, after the seminal work of DUSt3R \citep{wang2024dust3r}, a new paradigm, which embeds the entire workflow into a single feed-forward model, was created. DUSt3R and its successors (e.g., MASt3R \citep{leroy2024grounding}, VGGT \citep{wang2025vggt}) employ transformer-based architectures \citep{vaswani2017attention} to jointly regress camera poses and dense 3D geometry from an arbitrary set of images in a single forward pass. After this groundbreaking paradigm shift in 3D reconstruction, there has been an explosion of follow-up work \citep{tang2025mv,yang2025fast3r,zhang2025advances,zhang2025review}. 

\textbf{Uncertainty Quantification in Deep Vision Models.} Neural networks often produce some notion of confidence or uncertainty. In Bayesian modeling, uncertainty is usually decomposed into aleatoric (data) and epistemic (model) uncertainty \citep{der2009aleatory}. Aleatoric uncertainty captures inherent noise in the observations (e.g., sensor noise, low-texture regions) and cannot be reduced by more data, whereas epistemic uncertainty reflects the model’s lack of knowledge and can be mitigated with more data \citep{kendall2017uncertainties,gawlikowski2023survey}. This distinction is critical since \cite{wang2025vggt} state that VGGT outputs aleatoric uncertainty maps only, and does not account for epistemic uncertainty. We will discuss VGGT’s uncertainty modeling in detail in Section~\ref{sec:basics}. Unlike epistemic uncertainty, which typically requires multiple forward passes (e.g., via Monte Carlo dropout), aleatoric uncertainty can be learned directly from the data in a single forward pass. While these can be incorporated by various techniques like Bayesian Neural Networks \citep{gal2016dropout,neal2012bayesian} or Ensembling \citep{lakshminarayanan2017simple,ganaie2022ensemble}, they usually introduce high computational costs, lack rigorous theoretical analysis, or require careful design choices and modifications to the training process \citep{he2023survey}. 

\section{Basics of VGGT and its Uncertainty Estimation}\label{sec:basics}

VGGT \citep{wang2025vggt} processes an arbitrary number of input images from a scene using a transformer backbone. It is designed as a simple, pure learning-based model, i.e., without any rule-based assumptions like camera models or epipolar constraints. The architecture alternates between frame-wise and global self-attention layers, allowing the model to capture both local and global geometric relationships across views. Each image is tokenized using a pretrained DINO encoder \citep{oquab2024dinov2}, and special tokens are introduced to predict camera intrinsics and extrinsics. The outputs of VGGT are:

\begin{itemize}[nosep]
    \item exterior orientation and field of view for each image,
    \item depth maps per image and corresponding confidence maps,
    \item point maps in a global coordinate frame, i.e., one predicted 3D coordinate per pixel, and corresponding confidence maps,
    \item and dense feature maps for point tracking.
\end{itemize}

The model is trained to predict all these quantities jointly, by minimizing a weighted sum of four loss components, i.e., one loss component for each output type. This formulation is mathematically redundant (e.g., point maps can be derived from depth maps and camera parameters). This redundancy, however, improves performance through multi-task learning \citep{wang2025vggt}. In this paper, we focus on the 3D reconstruction, and hence on the depth maps and point maps as well as their corresponding confidence maps. 

\paragraph{Depth Map and Point Map Loss of VGGT.}

The depth loss
\begin{equation}\label{eq:depth:loss}
    \begin{aligned}
        \mathcal{L}_{\text{depth}} &= \left\| \boldsymbol{\Sigma}_D \odot (\boldsymbol{\hat{D}} - \boldsymbol{D}) \right\| \\
        & \quad + \left\| \boldsymbol{\Sigma}_D \odot (\nabla \boldsymbol{\hat{D}} - \nabla \boldsymbol{D}) \right\|
        - \alpha \log \boldsymbol{\Sigma}_D
    \end{aligned}
\end{equation}
is adapted from DUSt3R \citep{wang2024dust3r} and includes both residual and gradient terms, weighted by the predicted uncertainty map $\boldsymbol{\Sigma_D} \in \mathbb{R}^{H \times W}$,  where H and W are the image height and width. In Equation~(\ref{eq:depth:loss}), $\boldsymbol{\hat{D}}$ is the predicted depth map, $\boldsymbol{D}$ the ground truth depth map,  $\nabla$ the spatial gradient operator, $\odot$ the element-wise product, and $\alpha$ a regularization weight.

The point map loss 
\begin{equation}
    \begin{aligned}
        \mathcal{L}_{\text{pmap}} &= \left\| \boldsymbol{\Sigma_P} \odot (\boldsymbol{\hat{P}} - \boldsymbol{P}) \right\|\\ & \quad + \left\| \boldsymbol{\Sigma_P} \odot (\nabla \boldsymbol{\hat{P}} - \nabla \boldsymbol{P}) \right\| - \alpha \log \boldsymbol{\Sigma_P}
    \end{aligned}
\end{equation}
is defined analogously, where $\boldsymbol{\hat{P}}$ is the predicted 3D point map, $\boldsymbol{P}$ the ground truth point map, and $\boldsymbol{\Sigma_P} \in \mathbb{R}^{H \times W}$ is the predicted uncertainty map of the point maps.

In these loss formulations, the residuals  are weighted by the predicted uncertainty $\boldsymbol{\Sigma_D}$ or $\boldsymbol{\Sigma_P}$, meaning the model learns to downweight regions with large errors. The gradient terms enforce local smoothness and consistency in the spatial structure. The log penalty on the uncertainty prevents the model from trivially inflating $\boldsymbol{\Sigma_D}$ or $\boldsymbol{\Sigma_P}$ to reduce the residuals. This formulation is not probabilistic in the strict sense (i.e., not derived from a likelihood), but rather a heuristic uncertainty-aware loss that balances accuracy and confidence.

This mechanism allows VGGT to express spatially varying reliability in its predictions, which is particularly relevant for photogrammetric applications where geometric accuracy and error quantification play an important role. These uncertainties are not post-hoc estimates but are intrinsic outputs of the model, making them potentially suitable for downstream tasks such as uncertainty-aware outlier rejection, multi-view fusion, or meshing, as well as error propagation analysis.

\paragraph{Confidence and Uncertainty in VGGT.}
In the supplementary material of VGGT, \cite{wang2025vggt} describe that VGGT implements per-pixel \emph{aleatoric uncertainty} maps $\Sigma_D$ and $\Sigma_P$. The term \emph{uncertainty} usually means that large values express high uncertainty and vice versa. In contrast, the official implementation actually predicts per-pixel \emph{confidence} maps, where larger values represent higher confidence (lower uncertainty). Concretely, the weights $\boldsymbol{\Sigma} \in \{\boldsymbol{\Sigma_D}, \boldsymbol{\Sigma_P}\}$ which are passed through an \texttt{expp1} activation, yield confidence values
\begin{equation}
    \boldsymbol{C} \;=\; \mathrm{expp1}(\boldsymbol{\Sigma}) \;=\; e^{\boldsymbol{\Sigma}} + 1\,.
\end{equation}
Because of this activation, the values of the confidence maps are always larger than 1. However, this means that the confidence maps can contain values that represent negative weights $\boldsymbol{\Sigma}$, which are not meaningful in this context. Conversely, meaningful weights have a value larger than 2 in the confidence maps. We investigate this observation empirically in our experiments.

For analyses that conceptually operate on uncertainty, we convert confidence into a monotonic, scale-free uncertainty proxy $\boldsymbol{U}$, with
\begin{equation}\label{eq:uncertainty}
    \boldsymbol{U} \;=\; -\log (\boldsymbol{C} - 1) \, .
\end{equation}
We filter raw predictions in the confidence space (e.g., keep predictions with ${\boldsymbol{C}\;>\;2.0}$), but compute uncertainty metrics (PAvPU, pAC, pUI, sparsification, AUSE) in the uncertainty space using $\boldsymbol{U}$. 

\section{Evaluation Methodology}\label{sec:setup}

This section describes how we applied VGGT, provides relevant details about the DTU dataset and evaluation, and specifies the metrics used to evaluate point cloud and uncertainty quality.

\paragraph{VGGT.} We consistently evaluate the variant referred to as \textit{Feed-Forward} in \cite{wang2025vggt}, i.e., the raw outputs of VGGT without parameter adjustments, fine-tuning, or post-processing steps. This means in particular that we do not perform bundle adjustment in the post-processing. The reason for this is that the uncertainties can only be meaningfully evaluated for the feed-forward variant of VGGT, since they are not propagated by the integrated bundle adjustment with VGGSfM \citep{wang2024vggsfm}. We use the pretrained VGGT-1B checkpoint with 1.26 B parameters. By default, VGGT resizes all images to a size of $518 \times 518$ pixels and uses centered zero padding, i.e., padded pixels at the edges of the images are black. The points in the point maps refer to a coordinate system that is defined by the first image, so that these points can be evaluated directly as a point cloud. To compute point clouds from the depth maps, the estimated exterior orientation and field of view is used. In the following, reconstruction results obtained from the point map branch are referred to as \textit{VGGT-p}, while results from the depth map branch are denoted as \textit{VGGT-d}.

\paragraph{DTU Dataset.}
The DTU dataset \citep{aanaes2016large} is an established benchmark for evaluating MVS and 3D reconstruction algorithms. It was specifically designed to provide high-quality ground truth geometry under controlled conditions, making it particularly relevant for photogrammetric evaluation. It consists of 80 indoor scenes captured using a robotic arm equipped with a high-resolution camera. Each scene is captured from 49 or 64 viewpoints arranged on a hemispherical grid. For each scene, the dataset provides:
\begin{itemize}[nosep]
    \item High-resolution RGB images,
    \item Interior and exterior orientation,
    \item Structured-light-based reference geometry (ground truth),
    \item Surface normals and visibility masks.
\end{itemize}
The exterior orientation has been determined with a calibration board and resection in space with high accuracy. The scenes are illuminated under seven multiple lighting conditions (L1 - L7) to enable testing the robustness against photometric variation. Typically, light settings L3 or L7 are used for evaluations. We select L3 because it results in less overexposure in the images. The reference geometry was acquired using a structured light scanner with an average precision of approximately 0.14\,mm for the surface points, which corresponds to approximately 0.6 pixels, allowing for precise evaluation of reconstructed point clouds and depth maps. The scenes include both Lambertian and non-Lambertian surfaces, as well as varying levels of geometric complexity and occlusion.

\begin{figure*}[ht!]
    \centering
    \begin{subfigure}[t]{0.24\textwidth}
        \includegraphics[width=\textwidth]{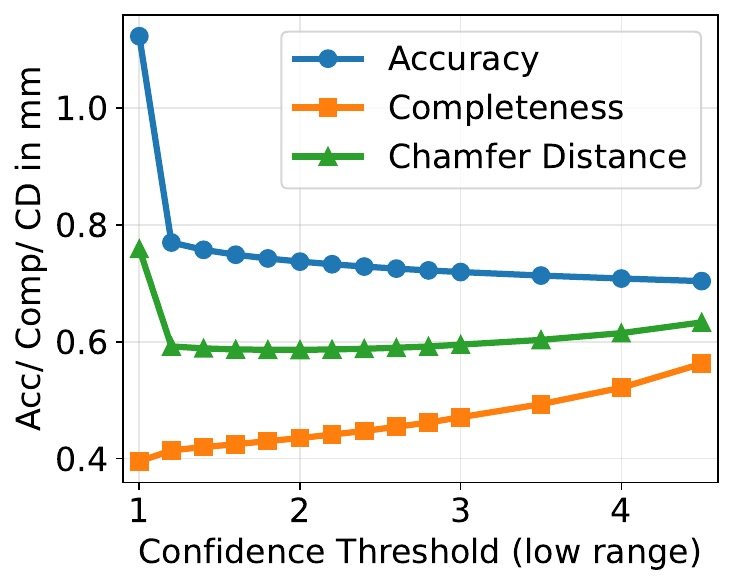}
    \end{subfigure}
    \hfill
    \begin{subfigure}[t]{0.24\textwidth}
        \includegraphics[width=\textwidth]{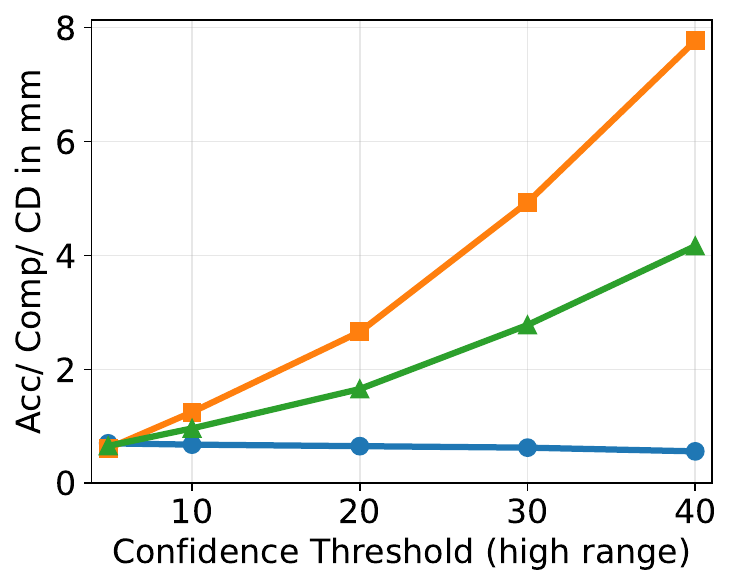}
    \end{subfigure}
    \hfill
    \begin{subfigure}[t]{0.24\textwidth}
        \includegraphics[width=\textwidth]{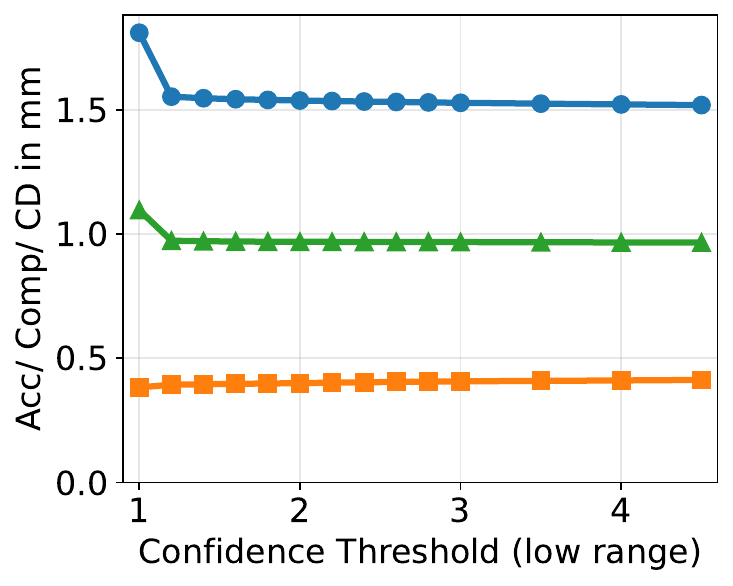}
    \end{subfigure}
    \hfill
    \begin{subfigure}[t]{0.24\textwidth}
        \includegraphics[width=\textwidth]{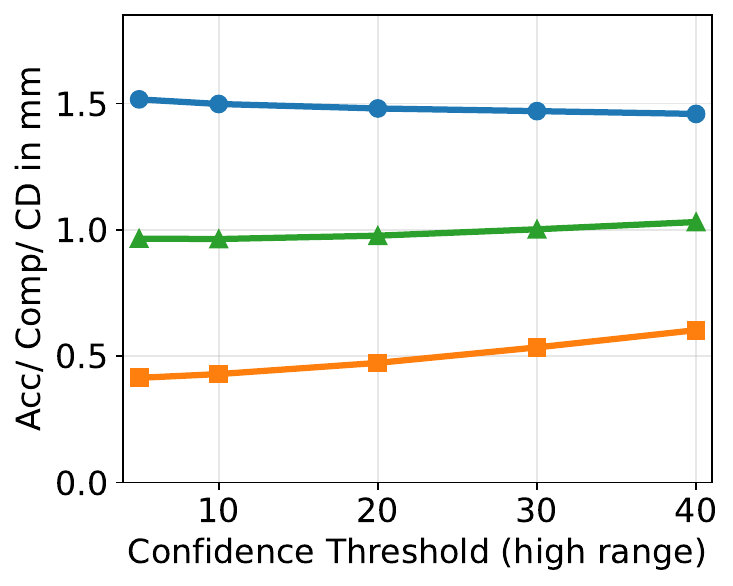}
    \end{subfigure}
    \caption{Accuracy, Completeness and Chamfer Distance (in mm) $\downarrow$ of VGGT-p (left) and VGGT-d (right) on the DTU dataset depending on the selected confidence threshold for filtering the raw output.}
    \label{fig:conf_thresh}
\end{figure*}

\paragraph{DTU Evaluation.}

The DTU benchmark provides a standardized evaluation protocol. The evaluation metrics are 
\begin{equation} \label{eq:acc}
     \text{Accuracy} = \frac{1}{|\boldsymbol{\mathcal{P}_r}|} \sum_{\boldsymbol{p}_r \in \boldsymbol{\mathcal{P}_r}} \min_{\boldsymbol{p_g} \in \boldsymbol{\mathcal{P}_g}} \| \boldsymbol{p_r} - \boldsymbol{p_g} \|_2\mathrm{,}
\end{equation}
i.e., the mean Euclidean distance from the points $\boldsymbol{p_r}$ in the reconstructed point cloud $\boldsymbol{\mathcal{P}_r}$ to the closest points $\boldsymbol{p_g}$ in the ground truth point cloud $\boldsymbol{\mathcal{P}_g}$,
\begin{equation}\label{eq:comp}
    \text{Completeness} = \frac{1}{|\boldsymbol{\mathcal{P}_g}|} \sum_{\boldsymbol{p_g} \in \boldsymbol{\mathcal{P}_g}} \min_{\boldsymbol{p_r} \in \boldsymbol{\mathcal{P}_r}} \| \boldsymbol{p_g} - \boldsymbol{p_r} \|_2\mathrm{,}
\end{equation}
i.e., the mean Euclidean distance from ground truth points to the closest reconstructed point, and the overall score, i.e., the average of Accuracy and Completeness, which is also known as Chamfer Distance. Note that for all three metrics, lower values indicate better performance ($\downarrow$). For the terms \enquote{Accuracy} and \enquote{Completeness}, this differs from the usual interpretation. If the metrics from Equations~(\ref{eq:acc}) and~(\ref{eq:comp}) are explicitly meant, we use the capitalized terms throughout the paper.

To compute reasonable distances, the reconstructed point clouds need to be registered to the ground truth. Thus, we estimate a coarse 3D similarity transformation based on the bounding boxes which is refined with the Iterative Closest Point algorithm by minimizing the Accuracy. The correctness of the registration is verified manually for all point clouds.

In the DTU evaluation, only points within a predefined evaluation mask are considered, ensuring that only visible and relevant regions are evaluated. Additionally, the point clouds are downsampled to a voxel size of 0.2\,mm, which corresponds to a conservative estimate of the accuracy of the ground truth \citep{aanaes2016large}.  Following \cite{yariv2020multiview} and a series of subsequent works like from \cite{huang20242dgs} or \cite{chen2024pgsr}, we evaluate on a fixed subset of 15 scenes.

\paragraph{Uncertainty Evaluation.}

To assess the uncertainty quality of VGGT, we employ the Patch Accuracy vs. Patch Uncertainty (PAvPU) metric \citep{mukhoti2018evaluating}
\begin{equation}
    \text{PAvPU} = \frac{(AC + IU)}{(AC + AU + IC + IU)}\enspace ,
\end{equation}
where $AC$, $AU$, $IC$, and $IU$ denote the number of accurate–certain, accurate–uncertain, inaccurate–certain, and inaccurate–uncertain predictions, respectively. PAvPU is based on two intuitive desiderata: (1) if a model is certain (confident) about its prediction, it should be accurate, and (2) if a model is uncertain, it may or may not be accurate \citep{mukhoti2018evaluating}. To determine whether a point is accurate or inaccurate and certain or uncertain, two thresholds $\tau_a$ and $\tau_u$ are necessary. We consider a point as accurate, when its Accuracy is smaller than an intuitive threshold of $\tau_a=1$\,mm. Following \cite{landgraf2025comparative}, we further consider a prediction as certain, when its uncertainty value is smaller than the median of the uncertainties ($<\tau_u=0.5$), and vice versa. \cite{mukhoti2018evaluating} additionally propose two intuitive probability metrics, that we also report: $\text{pAC} = \frac{AC}{AC + IC}$, i.e., the probability that the model is accurate given that the uncertainty is below $\tau_u$ and $\text{pUI} = \frac{IU}{IU + IC}$, i.e., the probability that the uncertainty of the model exceeds $\tau_u$ given that the prediction is inaccurate. Consequently, higher values for PAvPU, pAC, and pUI generally represent a higher uncertainty quality.



We further evaluate the uncertainty quality using Sparsification Curves and the Area Under the Sparsification Error (AUSE) \citep{ilg2018uncertainty}. The former visualize how well uncertainties correlate with actual prediction errors. They are constructed by sorting the predictions according to their uncertainty. Then, the most uncertain samples are gradually removed, while the remaining subset is used to compute the corresponding error metric (e.g., Chamfer Distance). If the uncertainty estimates perfectly reflect prediction errors, the error metric should decrease rapidly as uncertain samples are removed. The AUSE quantifies the deviation between the actual sparsification curve and an ideal curve, which represents the best possible ranking, which is obtained using ground-truth Accuracy instead of predicted uncertainty. It is calculated as the area between both curves, with lower AUSE values indicating higher uncertainty quality.

Finally, we present correlation curves. These represent the mean uncertainty across all points that lie within an Accuracy interval. The intuition behind the correlation curves is that there should tend to be a linear relationship between accuracy and uncertainty, such that the worse the accuracy, the higher the uncertainty should be. If this linear relationship exists and the uncertainty is calibrated, a metric statement about accuracy can be derived from the uncertainty prediction, which is a prerequisite for subsequent statistical evaluations such as hypothesis testing. Deviations from this intuition are represented by accurate but uncertain points, which are acceptable and should therefore have no negative impact on an uncertainty quality metric. Consequently, for points with an Accuracy better than $\tau_a$, this requirement does not apply strictly.

\section{Results of VGGT on the DTU-Dataset}\label{sec:Experiments}
This section presents the evaluation results of VGGT on the DTU dataset, with a particular focus on the uncertainty quality. Since uncertainty quality is best interpreted relative to the underlying prediction error, we first report the reconstruction performance using standard DTU evaluation metrics, as outlined in Section \ref{sec:setup}. Lastly, we analyze the correlation between VGGT’s prediction errors and its estimated uncertainties.



\subsection{Preliminary study on confidence filtering}

Filtering the raw output from VGGT based on confidence values has a significant impact on point cloud quality (see Fig.\@~\ref{fig:conf_filter}). Therefore, we first conduct a preliminary study to determine the best confidence threshold. 

Fig.\@~\ref{fig:conf_thresh} presents the results of the DTU evaluation in terms of Accuracy, Completeness, and Chamfer Distance averaged over all scenes, depending on the selected confidence threshold for filtering. For both VGGT-p and VVGT-d, the analysis shows that the higher the confidence threshold is set, the lower the Accuracy and the higher the Completeness. However, Completeness increases faster than Accuracy decreases. Very similar Chamfer Distances are achieved after filtering with confidence thresholds ranging from 1.2 to 5. However, if all points are included in the evaluation (threshold 1.0), the result is significantly worse. This clearly indicates that data points with the lowest confidences should be excluded from further analysis. 
For VGGT-d (Fig.\@~\ref{fig:conf_thresh}, right), filtering based on the confidence threshold has only a very minor effect. This is due to poor confidence predictions, which we will discuss again in Section \ref{sec:qual_unc}.
For VGGT-p (Fig.\@~\ref{fig:conf_thresh}, left), the optimal confidence threshold across the scenes is always between 1.2 and 3.5 with a mean value of 1.9 and a standard deviation of 0.6. Consequently, empirical evidence also shows that the causally meaningful (cf. Section~\ref{sec:basics}) threshold of 2.0 is a good choice. Therefore, further investigations are carried out based on the outputs filtered with a threshold of 2.0.

\subsection{Performance of VGGT on the DTU-Dataset}

Table \ref{tab:results_comparison} provides an overview of a number of commonly used, current methods for 3D reconstruction from multiple images. Some of these methods require known external orientations as input, which is usually calculated by SfM using COLMAP. The values are not exactly comparable, as different subsets of scenes were used for evaluation in some cases, and the runtimes were determined using different hardware. However, the table is intended to serve as a rough overview only. VGGT achieves mean Chamfer Distances of less than one millimeter for both branches, even though it was not trained on the DTU data and requires only a few seconds of computation time on an A100 GPU. Most other methods are considerably slower. Despite the longer runtimes, however, these methods hardly achieve a higher point cloud quality. One exception is MAST3R, which is fast and has a lower Chamfer Distance than both VGGT variants. However, ground truth poses were used to achieve this result \citep{wang2025vggt}.
For comparison, \cite{wang2025vggt} also report results for the DTU dataset. They achieve a lower Chamfer Distance of 0.38\,mm and report a runtime of 1.8 seconds with a more powerful graphics card than ours. We suspect that \cite{wang2025vggt} perform bundle adjustment in post-processing, thereby achieving a lower Chamfer Distance. However, this and other details of their experiment are not explicitly mentioned in the paper, so the experiments cannot be reproduced exactly.

\begin{table}[t]
  \centering
  \tablesqueeze
  \caption{Quantitative results of mean Chamfer Distance (CD) (in mm) $\downarrow$ across multiple scenes of the DTU dataset. The table is intended to serve as a rough overview. The values are not exactly comparable, as different subsets of scenes were used for evaluation in some cases, and the runtimes were determined using different hardware. We were unable to find runtimes for the DTU dataset for MASt3R, COLMAP, and DUSt3R in the literature, so we report a rough estimate based on the processing of 49 images. The results for the two VGGT variants are based on our evaluation with an A100 and measure the inference time of the model.}
  \label{tab:results_comparison}

  \begin{threeparttable}
  \begin{tabular}{
     l
     S[table-format=1.2] 
     l                  
     l
  }
    \toprule
    \textbf{Method} & \textbf{CD} & \textbf{Time}\\
    \midrule
    Known exterior orientation\\
    \midrule
    Neuralangelo~\citep{li2023neuralangelo}
      & 0.61 & {$>128$h} \\
    2DGS~\citep{huang20242dgs}
      & 0.80 & {$\sim$20min} \\
    PGSR~\citep{chen2024pgsr}
      & 0.52 & {$\sim$30min} \\
    MASt3R~\citep{leroy2024grounding}&0.37&{3-10min} \\
    \midrule
    Unknown exterior orientation\\
    \midrule
    COLMAP~\citep{schonberger2016pixelwise}&0.53&{1-2h}\\
    DUSt3R~\citep{wang2024dust3r}&1.74&{3-10min}\\    
    \addlinespace[3pt]
    VGGT-p~\citep{wang2025vggt} & 0.59 & {$\sim$ 5sec} \\
    VGGT-d~\citep{wang2025vggt} & 0.97 & {$\sim$ 5sec} \\
    \bottomrule
  \end{tabular}
  \end{threeparttable}
\end{table}

Table \ref{tab:vggt_results_per_scene} shows the results of VGGT on the DTU dataset in terms of Accuracy (Acc), Completeness (Comp), and Chamfer Distance (CD). VGGT-p, i.e., the point map branch, is significantly more accurate than VGGT-d on the DTU dataset. In contrast, VGGT-d is more complete on average. The quality of the results is very similar for the different scenes; no pattern can be discerned with this regard.

\begin{table*}[ht!]
  \centering
  \renewcommand{\arraystretch}{1.06}
  \setlength{\tabcolsep}{4.0pt}
  \caption{Accuracy (Acc), Completeness (Comp), and Chamfer Distance (CD) (in mm) per scene on the DTU dataset. VGGT-p denotes the point map branch and VGGT-d the depth map branch of VGGT.}
  \label{tab:vggt_results_per_scene}

  \begin{threeparttable}
      \begin{tabular}{
        l l
        S[table-format=1.2] S[table-format=1.2] S[table-format=1.2] S[table-format=1.2] S[table-format=1.2]
        S[table-format=1.2] S[table-format=1.2] S[table-format=1.2] S[table-format=1.2] S[table-format=1.2]
        S[table-format=1.2] S[table-format=1.2] S[table-format=1.2] S[table-format=1.2] S[table-format=1.2]
        S[table-format=1.2]
      }
        \toprule
        \textbf{Method} & \textbf{Metric}
          & \textbf{24} & \textbf{37} & \textbf{40} & \textbf{55} & \textbf{63}
          & \textbf{65} & \textbf{69} & \textbf{83} & \textbf{97} & \textbf{105}
          & \textbf{106} & \textbf{110} & \textbf{114} & \textbf{118} & \textbf{122}
          & \textbf{Mean} \\
        \midrule
        \multirow{3}{*}{\textbf{VGGT-p}}
          & acc $\downarrow$   & 0.86 & 0.95 & 0.97 & 0.68 & 0.82 & 0.56 & 0.85 & 0.76 & 0.75 & 0.69 & 0.58 & 0.63 & 0.71 & 0.61 & 0.65 & \textbf{0.74} \\
          & comp $\downarrow$ & 0.44 & 0.60 & 0.58 & 0.36 & 0.43 & 0.45 & 0.51 & 0.40 & 0.55 & 0.36 & 0.32 & 0.31 & 0.51 & 0.36 & 0.33 & 0.44 \\
          & CD $\downarrow$ & 0.65 & 0.77 & 0.78 & 0.52 & 0.62 & 0.50 & 0.68 & 0.58 & 0.65 & 0.53 & 0.45 & 0.47 & 0.61 & 0.49 & 0.49 & \textbf{0.59} \\
        \addlinespace[2pt] 
        \hdashline[1pt/2pt]
        \addlinespace[2pt]
        \multirow{3}{*}{\textbf{VGGT-d}}
          & Acc $\downarrow$
          & 1.24 & 1.55 & 1.85 & 1.55 & 2.03 & 1.61 & 1.38 & 1.71 & 1.71 & 1.50 & 1.28 & 1.39 & 1.11 & 1.38 & 1.65 & 1.53 \\
          & Comp $\downarrow$
          & 0.30 & 0.42 & 0.42 & 0.33 & 0.75 & 0.39 & 0.31 & 0.62 & 0.49 & 0.37 & 0.30 & 0.36 & 0.25 & 0.34 & 0.35 & \textbf{0.40} \\
          & CD $\downarrow$
          & 0.83 & 0.99 & 1.14 & 0.94 & 1.39 & 1.00 & 0.84 & 1.17 & 1.10 & 0.93 & 0.79 & 0.88 & 0.68 & 0.86 & 1.00 & 0.97 \\
        \bottomrule
      \end{tabular}
  \end{threeparttable}
\end{table*}

\begin{figure*}[ht!]
    \centering

    \begin{minipage}[c]{0.1\textwidth}
        \makebox[\textwidth]{}
    \end{minipage}%
    \begin{minipage}[c]{0.9\textwidth}
        \makebox[0.245\textwidth]{\centering \textbf{Prediction}}%
        \makebox[0.245\textwidth]{\centering \textbf{Pointwise Accuracy}}%
        \makebox[0.245\textwidth]{\centering \textbf{Uncertainty}}%
        \makebox[0.245\textwidth]{\centering \textbf{AC/ AU/ IC/ IU}}%
    \end{minipage}\\
    
    \begin{minipage}[c]{0.1\textwidth}
        \vspace{0.5cm}
        \centering \textbf{VGGT-p, \\ scene 24}
    \end{minipage}%
    \begin{minipage}[c]{0.9\textwidth}
        \begin{subfigure}[t]{0.245\textwidth}
            \includegraphics[width=\textwidth, trim=6cm 7cm 4cm 5cm, clip]{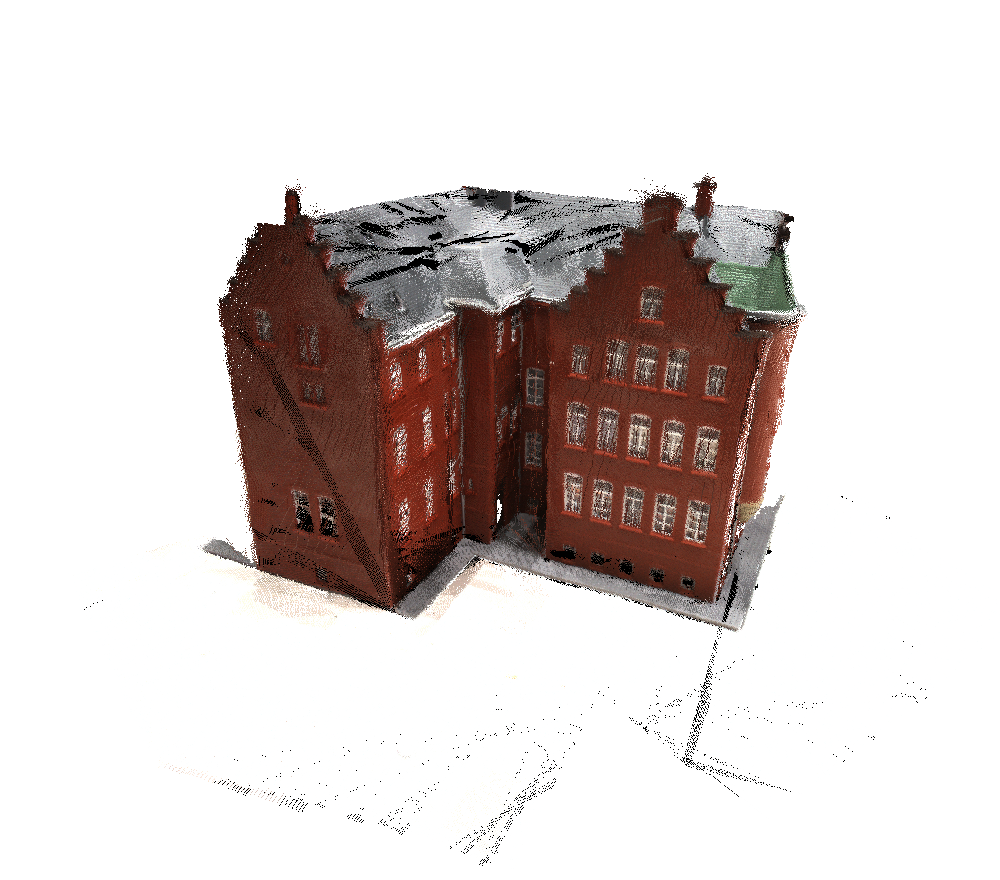}
        \end{subfigure}
        \begin{subfigure}[t]{0.245\textwidth}
            \includegraphics[width=\textwidth, trim=6cm 7cm 4cm 5cm, clip]{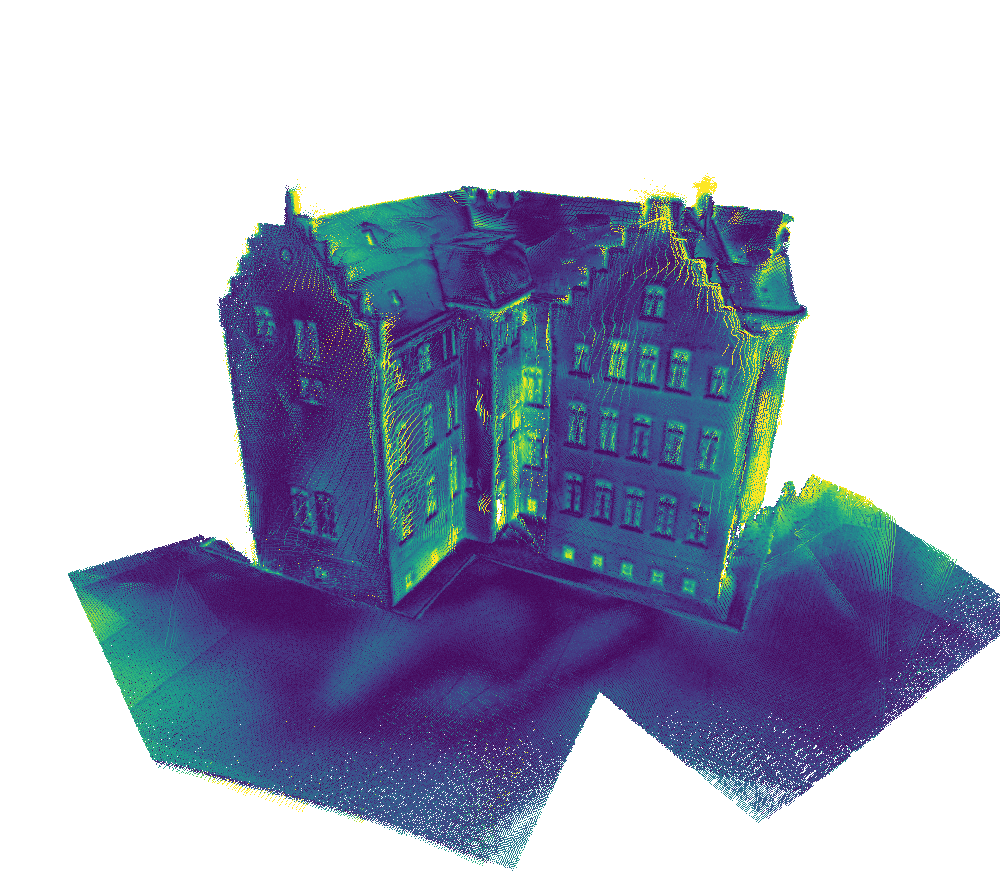}
        \end{subfigure}
        \begin{subfigure}[t]{0.245\textwidth}
            \includegraphics[width=\textwidth, trim=6cm 7cm 4cm 5cm, clip]{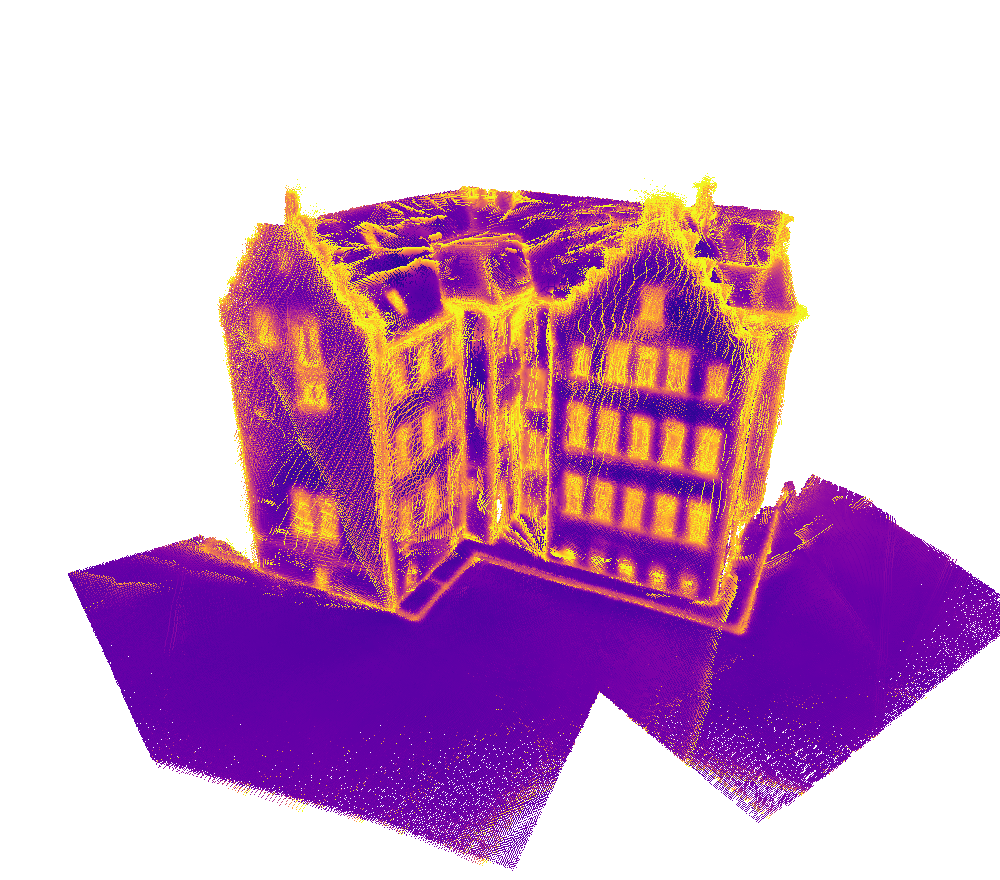}
        \end{subfigure}
        \begin{subfigure}[t]{0.245\textwidth}
            \includegraphics[width=\textwidth, trim=6cm 7cm 4cm 5cm, clip]{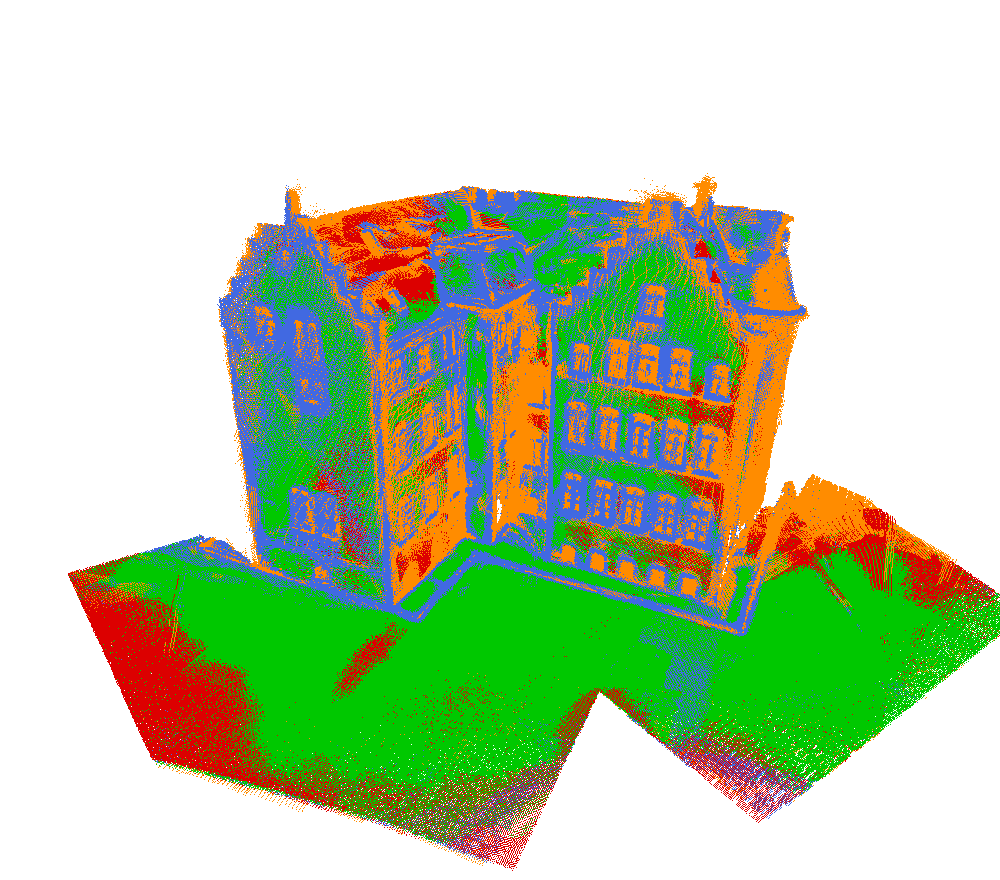}
        \end{subfigure}
    \end{minipage}\\[0.5ex]

    \begin{minipage}[c]{0.1\textwidth}
        \vspace{0.5cm}
        \centering \textbf{VGGT-d, \\ scene 24}
    \end{minipage}%
    \begin{minipage}[c]{0.89\textwidth}
        \begin{subfigure}[t]{0.245\textwidth}
            \includegraphics[width=\textwidth, trim=6cm 7cm 4cm 5cm, clip]{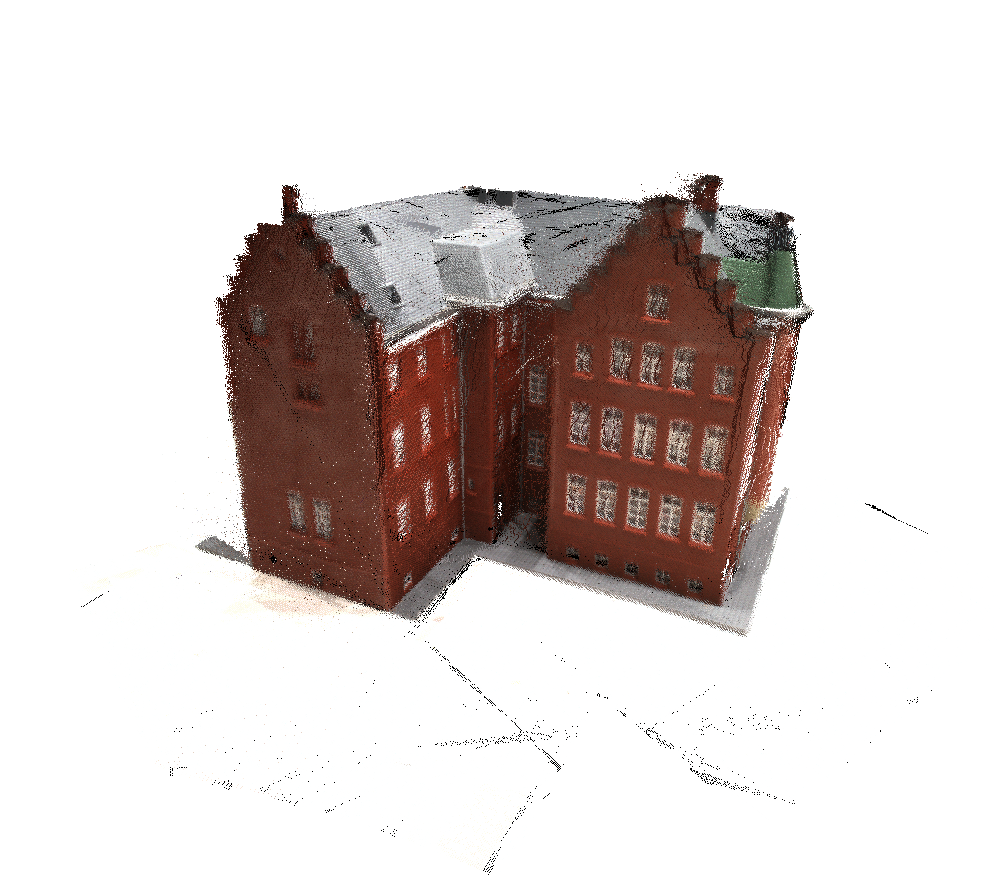}
        \end{subfigure}
        \begin{subfigure}[t]{0.245\textwidth}
            \includegraphics[width=\textwidth, trim=6cm 7cm 4cm 5cm, clip]{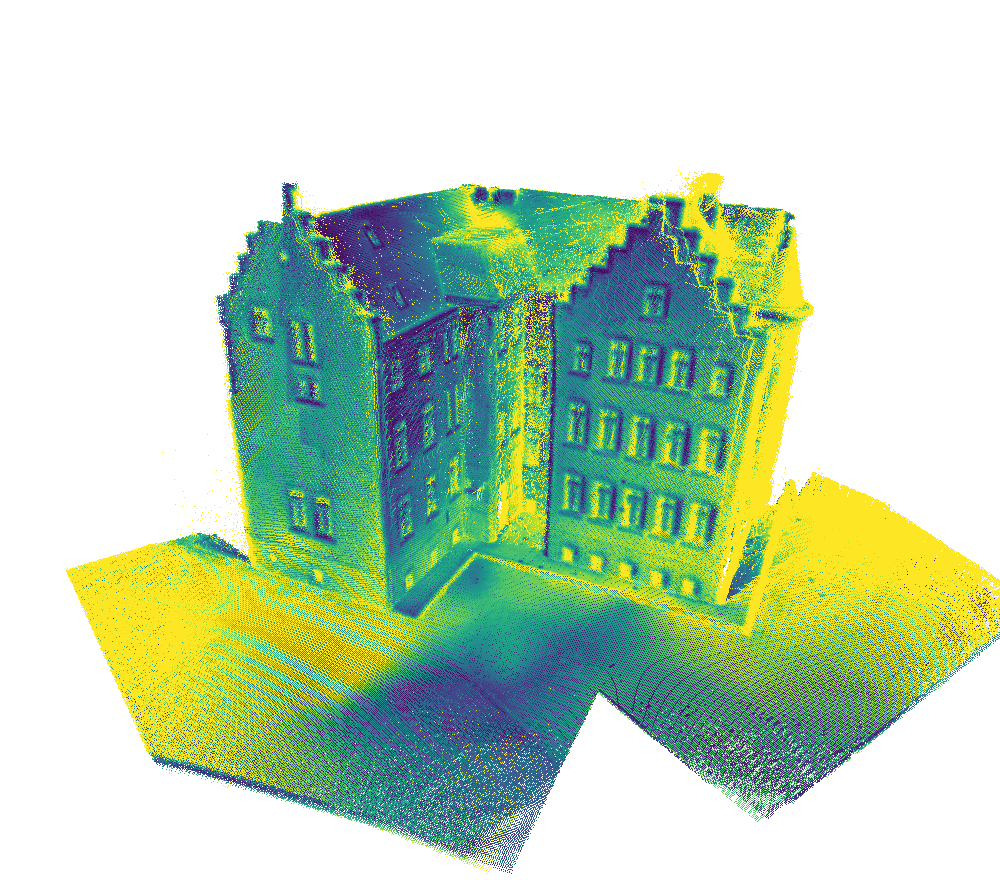}
        \end{subfigure}
        \begin{subfigure}[t]{0.245\textwidth}
            \includegraphics[width=\textwidth, trim=6cm 7cm 4cm 5cm, clip]{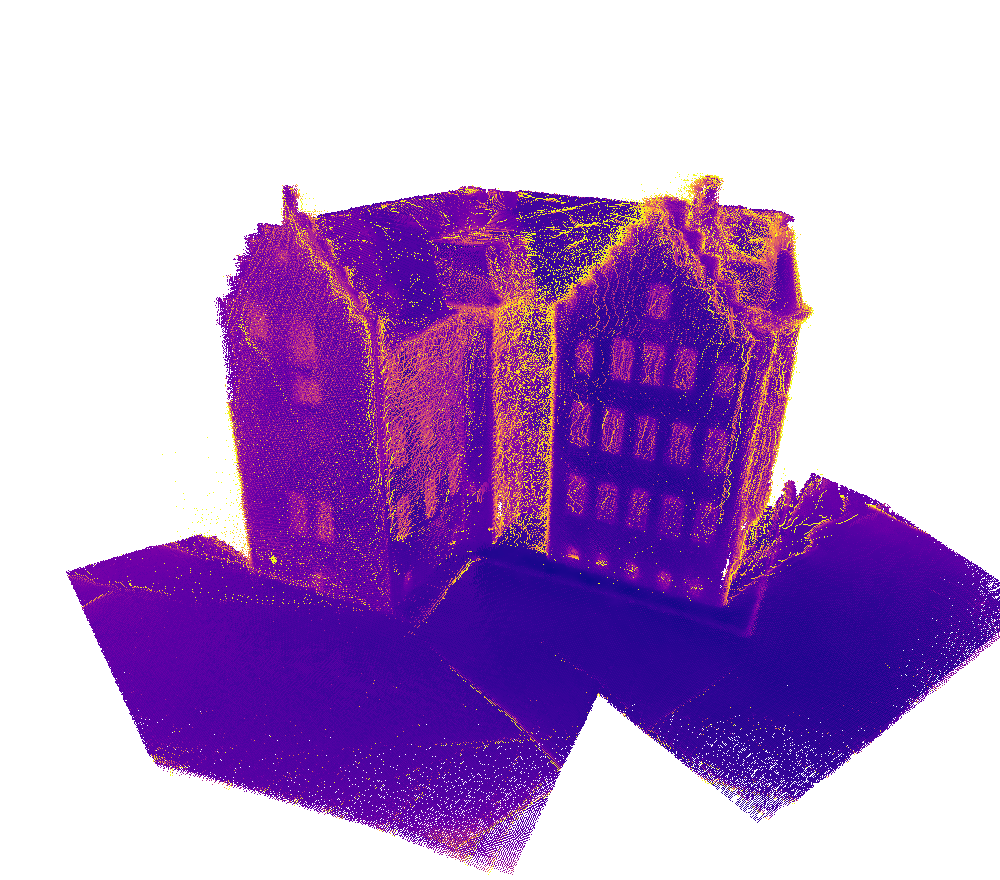}
        \end{subfigure}
        \begin{subfigure}[t]{0.245\textwidth}
            \includegraphics[width=\textwidth, trim=6cm 7cm 4cm 5cm, clip]{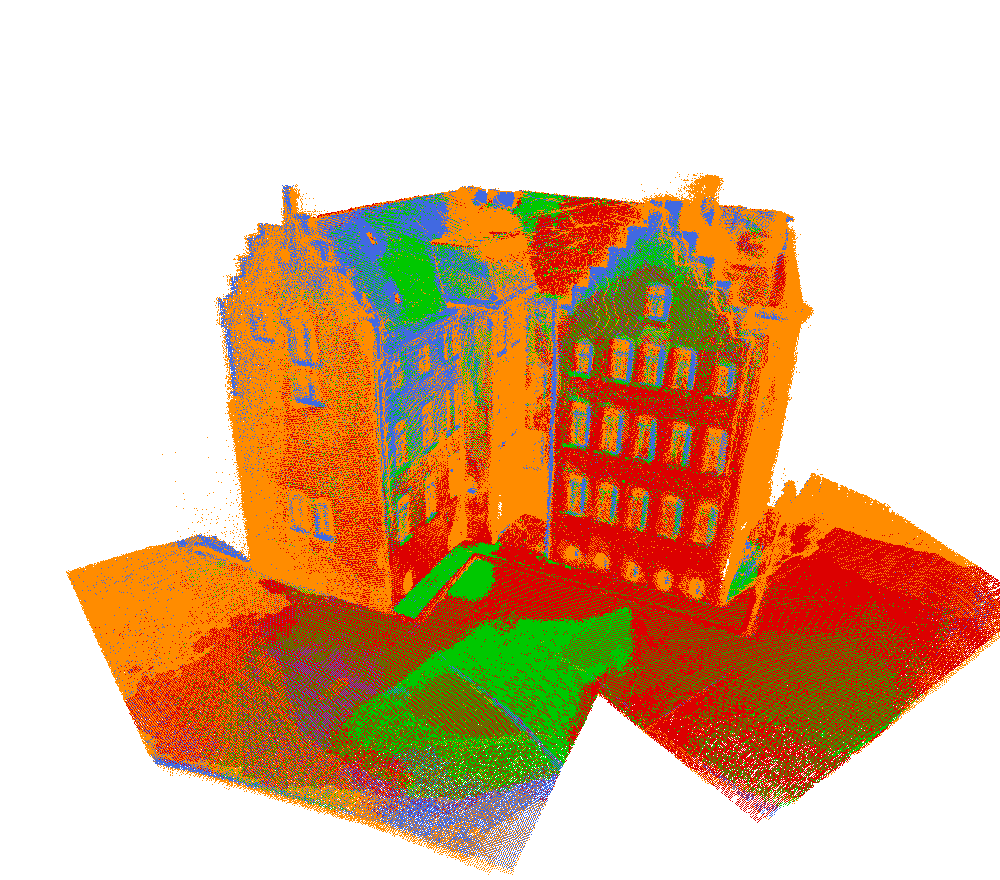}
        \end{subfigure}
    \end{minipage}\\[0.5ex]

    \begin{minipage}[c]{0.1\textwidth}
        \centering \textbf{VGGT-p, \\ scene 69}
    \end{minipage}%
    \begin{minipage}[c]{0.89\textwidth}
        \begin{subfigure}[t]{0.245\textwidth}
            \includegraphics[width=\textwidth, trim=4cm 8cm 4cm 2cm, clip]{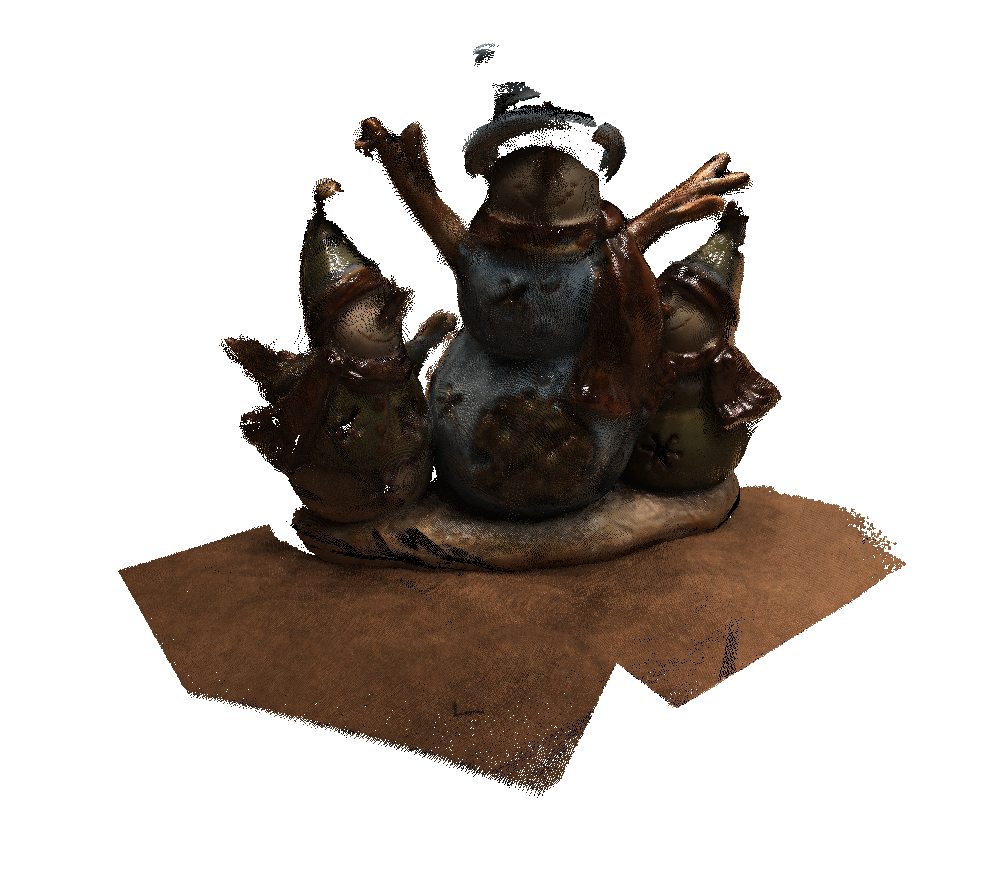}
        \end{subfigure}
        \begin{subfigure}[t]{0.245\textwidth}
            \includegraphics[width=\textwidth, trim=4cm 8cm 4cm 2cm, clip]{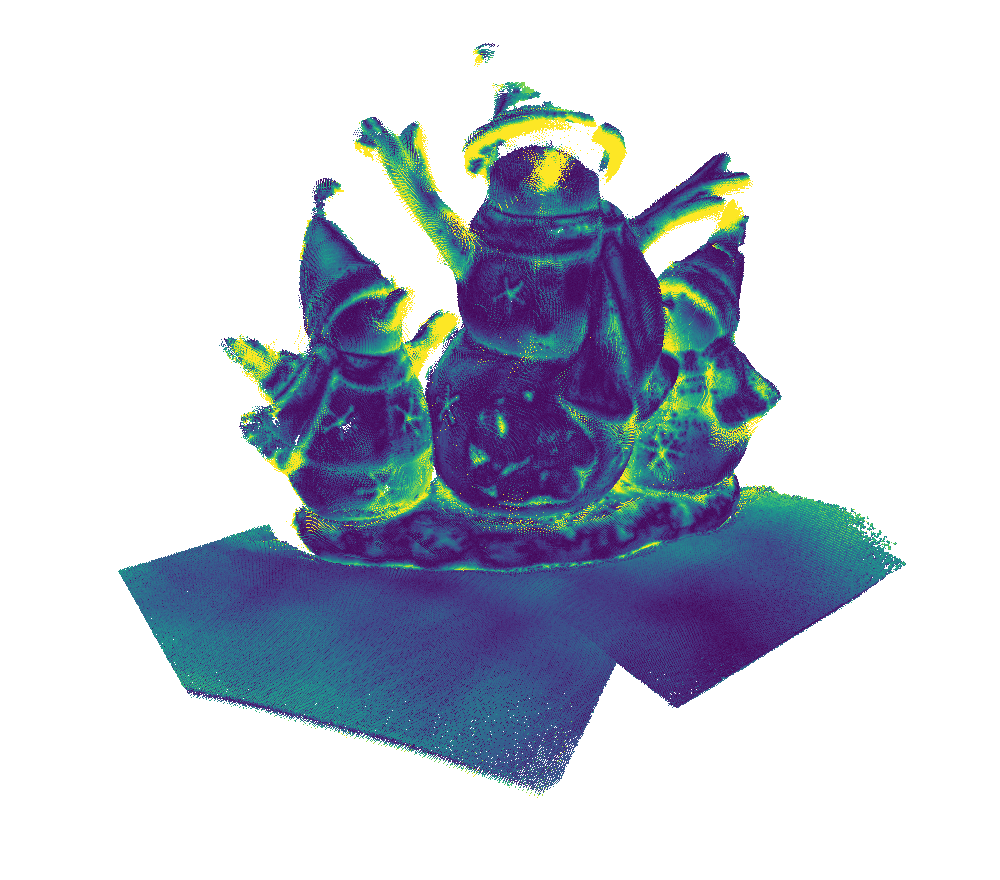}
        \end{subfigure}
        \begin{subfigure}[t]{0.245\textwidth}
            \includegraphics[width=\textwidth, trim=4cm 8cm 4cm 2cm, clip]{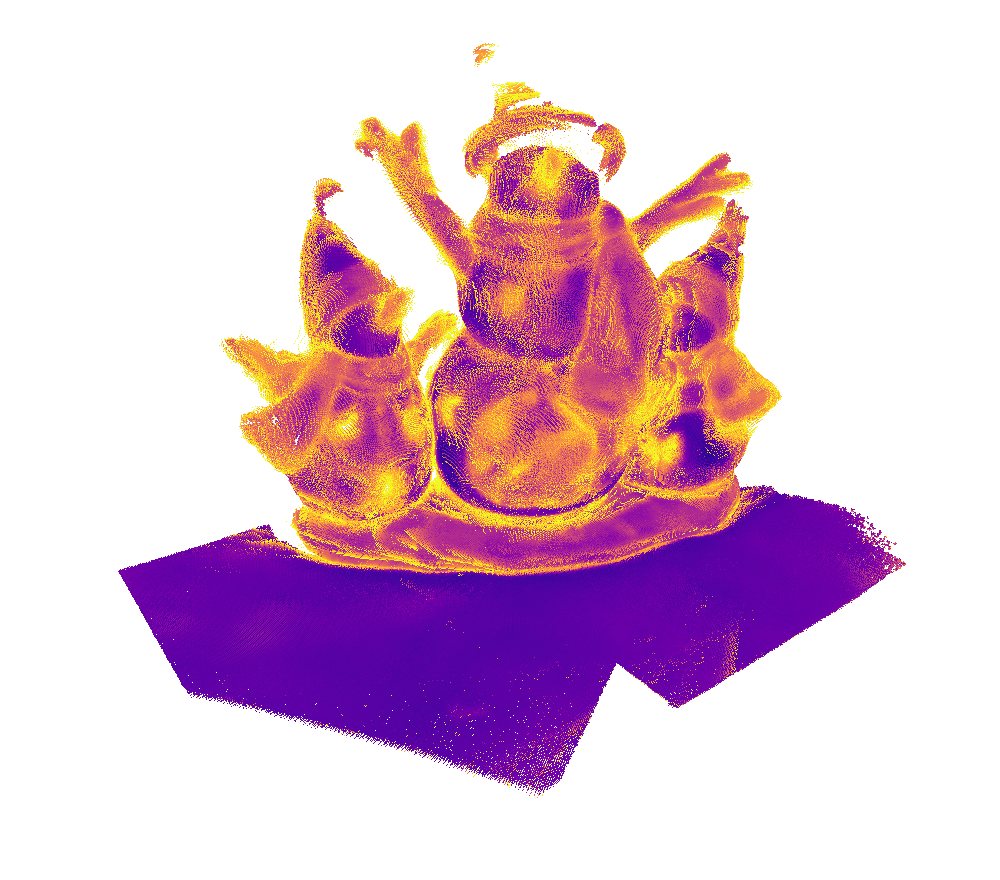}
        \end{subfigure}
        \begin{subfigure}[t]{0.245\textwidth}
            \includegraphics[width=\textwidth, trim=4cm 8cm 4cm 2cm, clip]{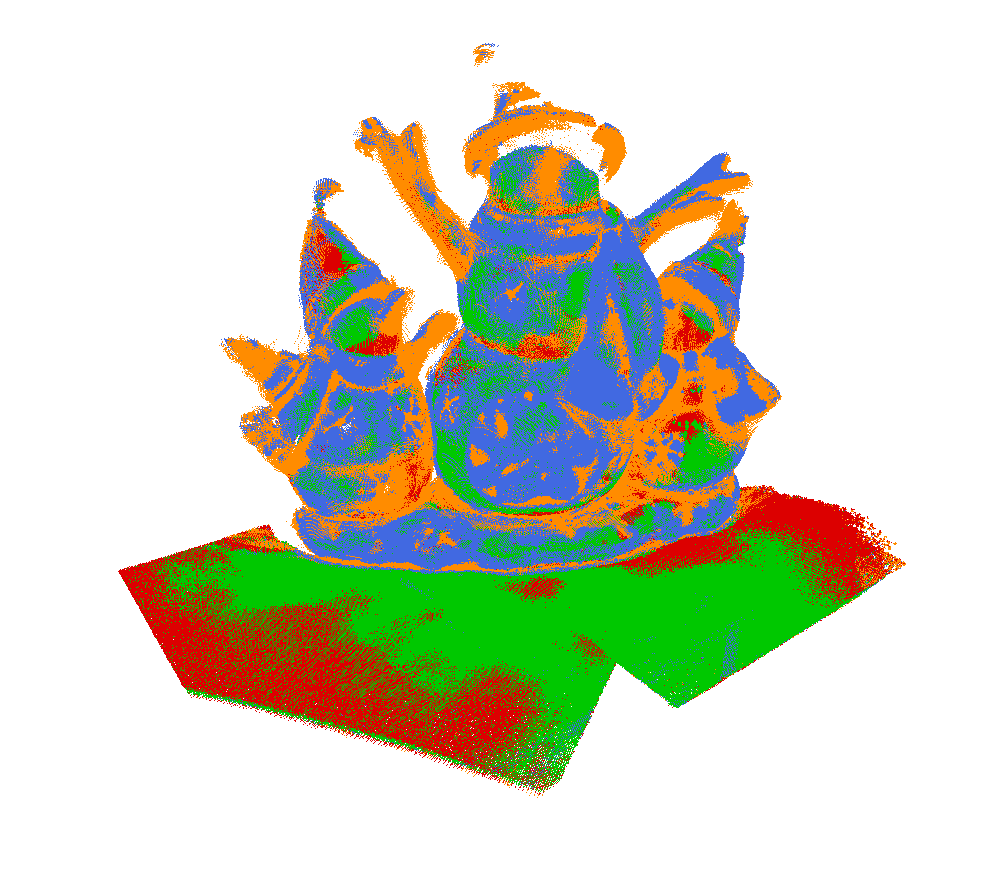}
        \end{subfigure}
    \end{minipage}\\[0.5ex]

    \begin{minipage}[c]{0.1\textwidth}
        \vspace{0.5cm}
        \centering \textbf{VGGT-d, \\ scene 69}
    \end{minipage}%
    \begin{minipage}[c]{0.89\textwidth}
        \begin{subfigure}[t]{0.245\textwidth}
            \includegraphics[width=\textwidth, trim=4cm 8cm 4cm 2cm, clip]{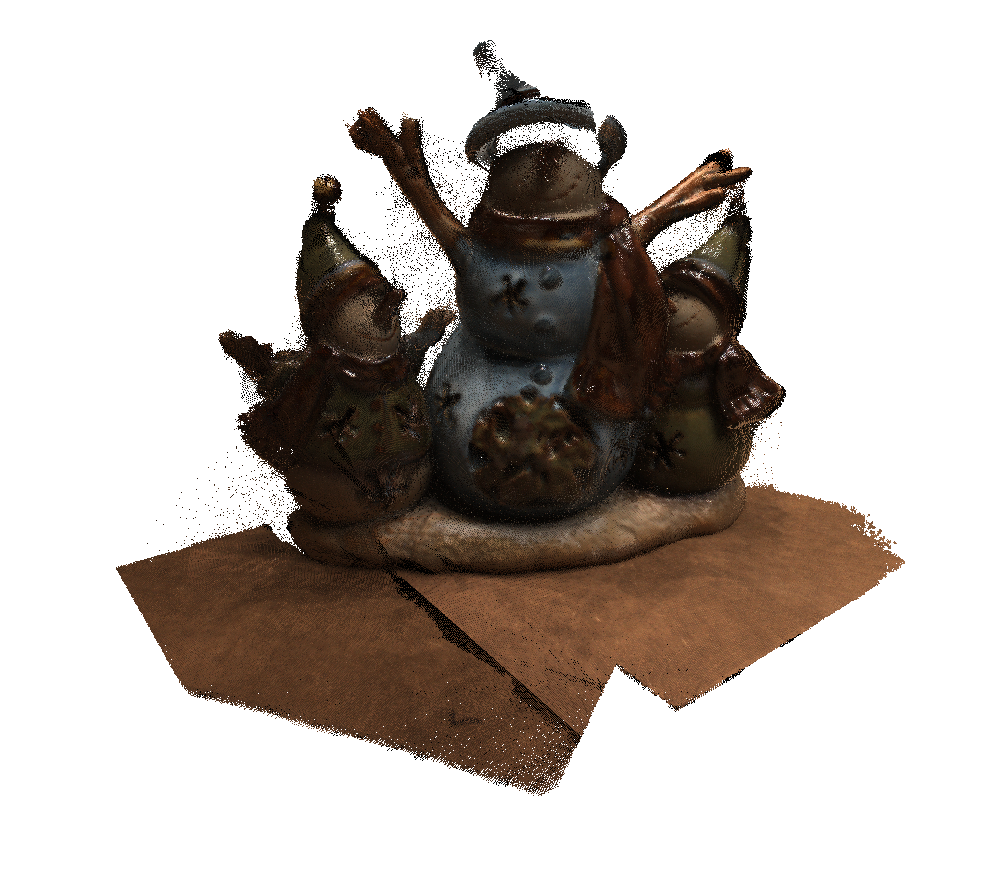}
        \end{subfigure}
        \begin{subfigure}[t]{0.245\textwidth}
            \includegraphics[width=\textwidth, trim=4cm 8cm 4cm 2cm, clip]{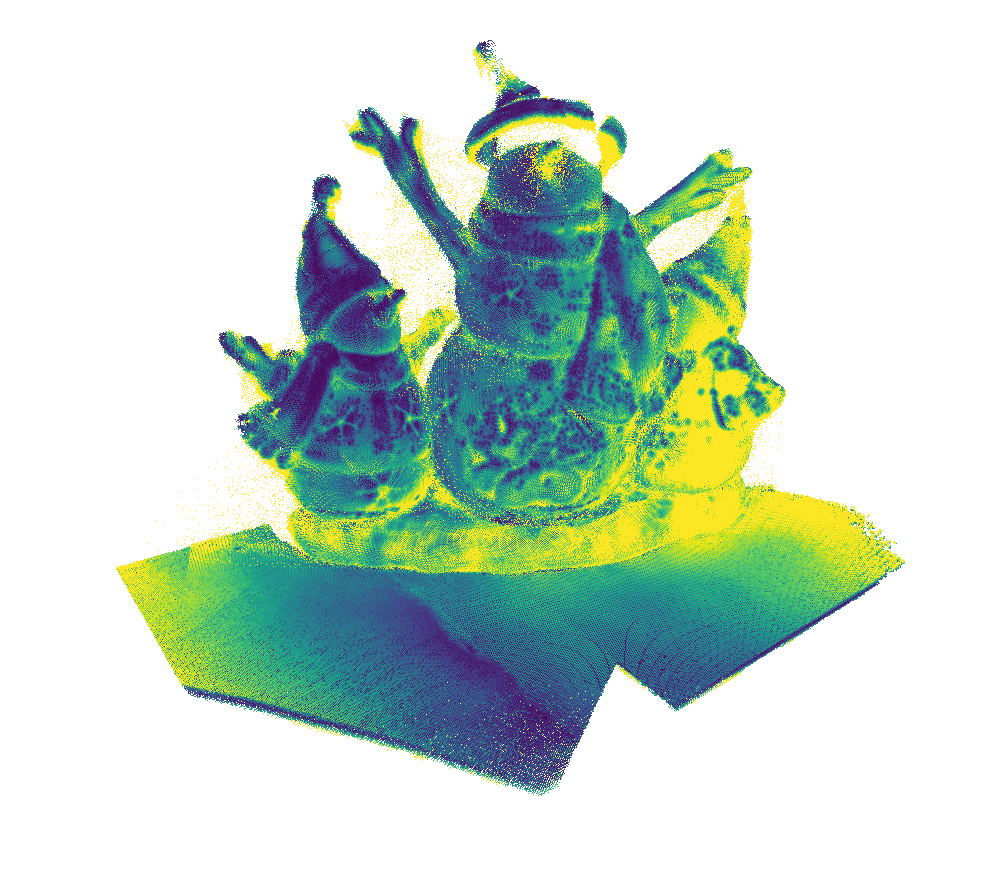}
        \end{subfigure}
        \begin{subfigure}[t]{0.245\textwidth}
            \includegraphics[width=\textwidth, trim=4cm 8cm 4cm 2cm, clip]{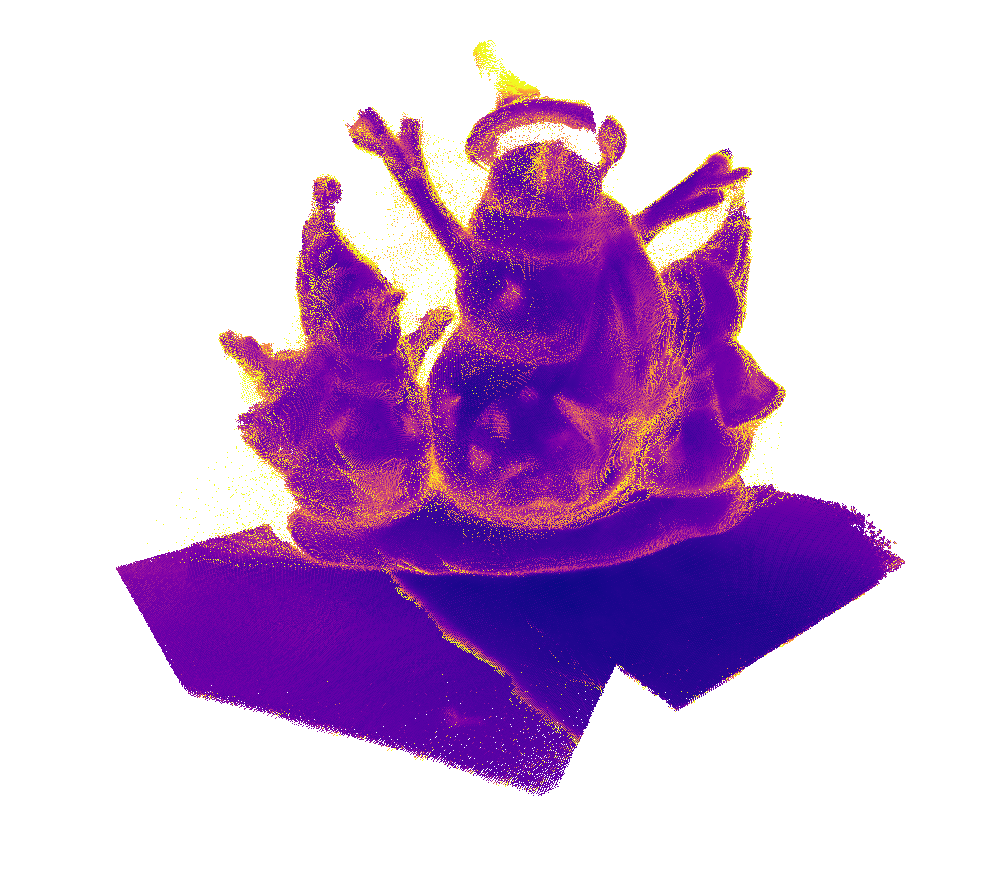}
        \end{subfigure}
        \begin{subfigure}[t]{0.245\textwidth}
            \includegraphics[width=\textwidth, trim=4cm 8cm 4cm 2cm, clip]{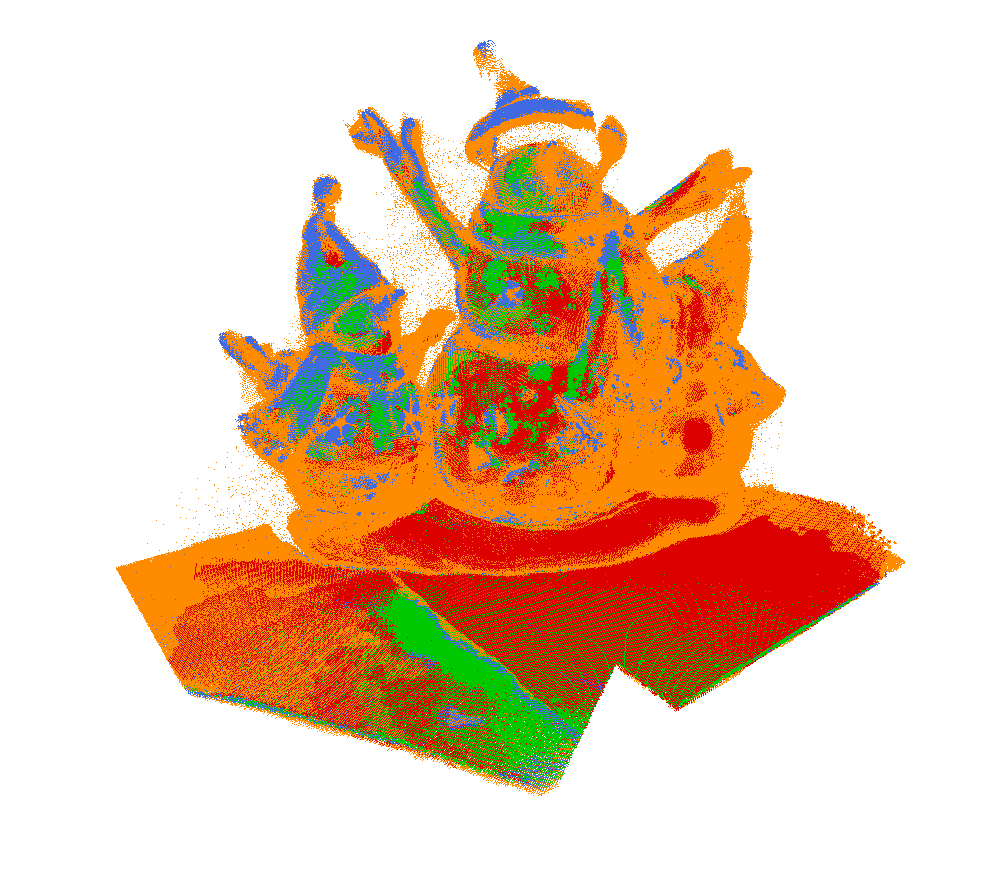}
        \end{subfigure}
    \end{minipage}\\[0.5ex]

    \begin{minipage}[c]{0.1\textwidth}
        \centering \textbf{VGGT-p, \\ scene 122}
    \end{minipage}%
    \begin{minipage}[c]{0.89\textwidth}
        \begin{subfigure}[t]{0.245\textwidth}
            \includegraphics[width=\textwidth, trim=5cm 8cm 4cm 3cm, clip]{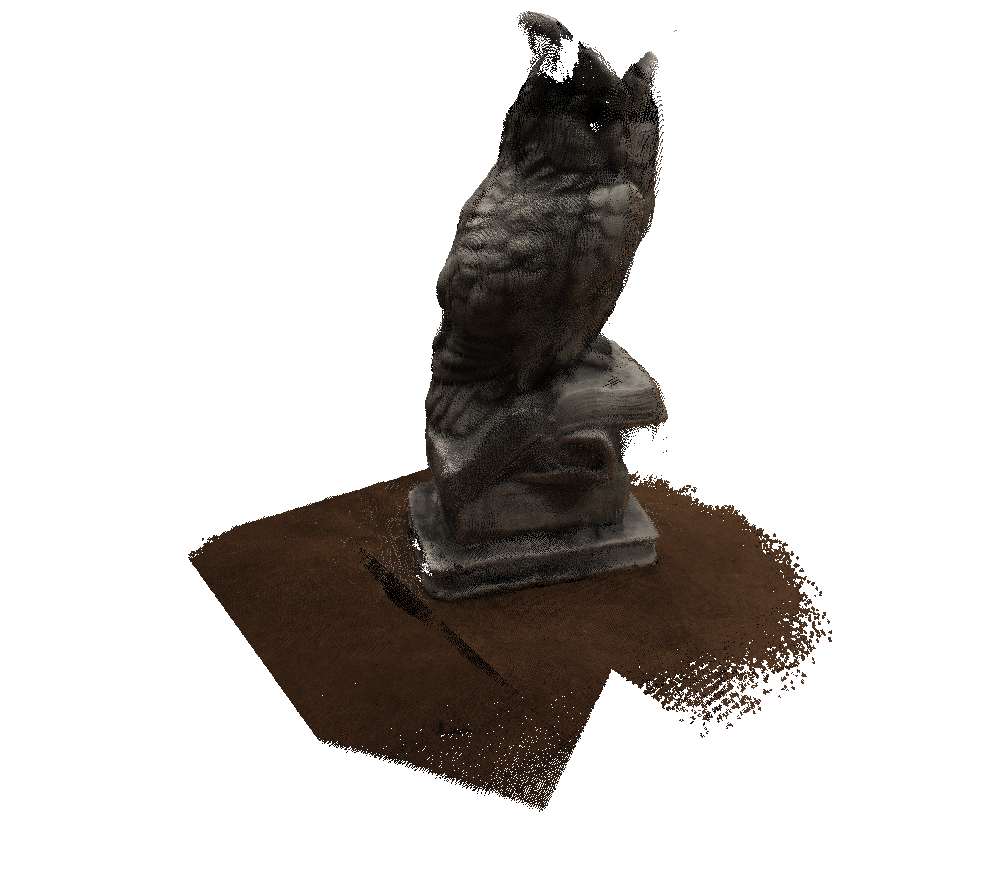}
        \end{subfigure}
        \begin{subfigure}[t]{0.245\textwidth}
            \includegraphics[width=\textwidth, trim=5cm 8cm 4cm 3cm, clip]{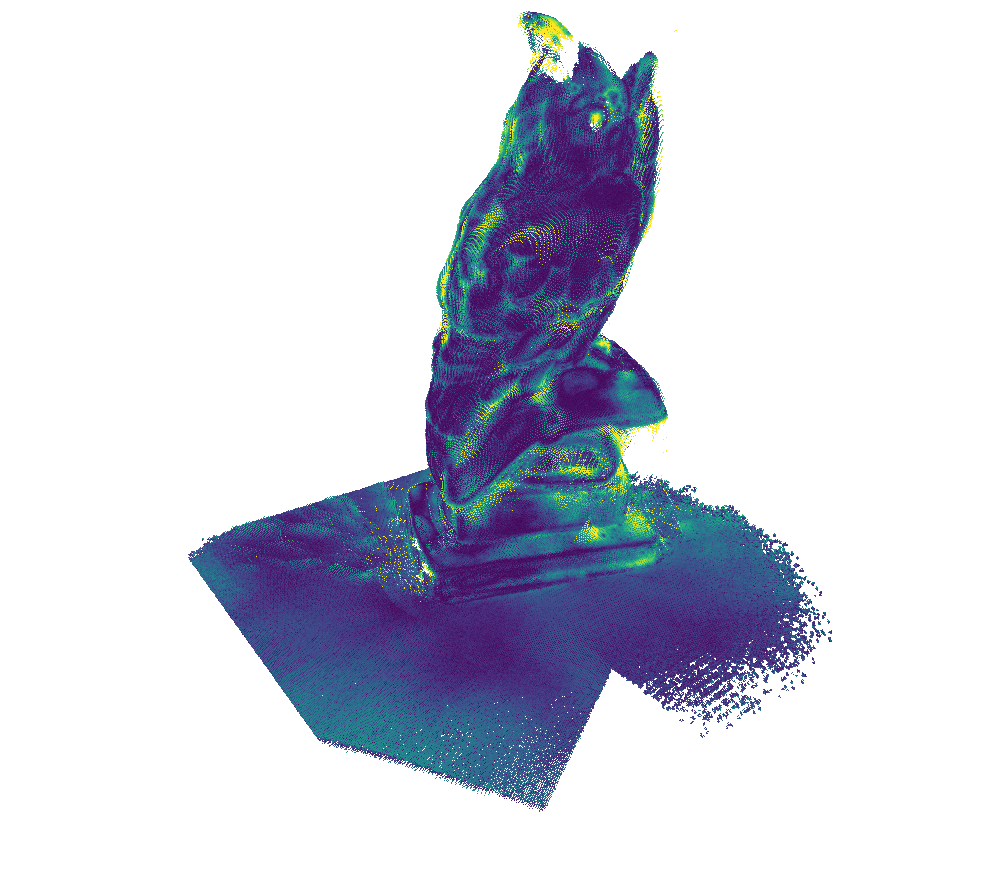}
        \end{subfigure}
        \begin{subfigure}[t]{0.245\textwidth}
            \includegraphics[width=\textwidth, trim=5cm 8cm 4cm 3cm, clip]{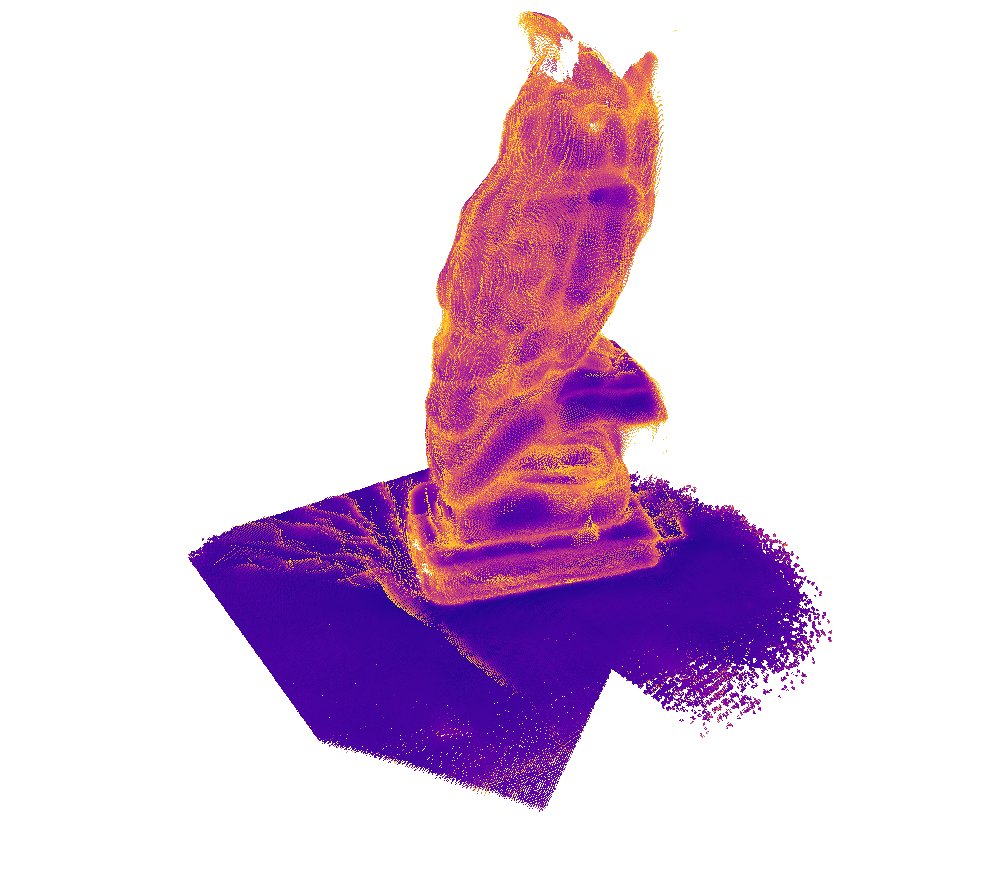}
        \end{subfigure}
        \begin{subfigure}[t]{0.245\textwidth}
            \includegraphics[width=\textwidth, trim=5cm 8cm 4cm 3cm, clip]{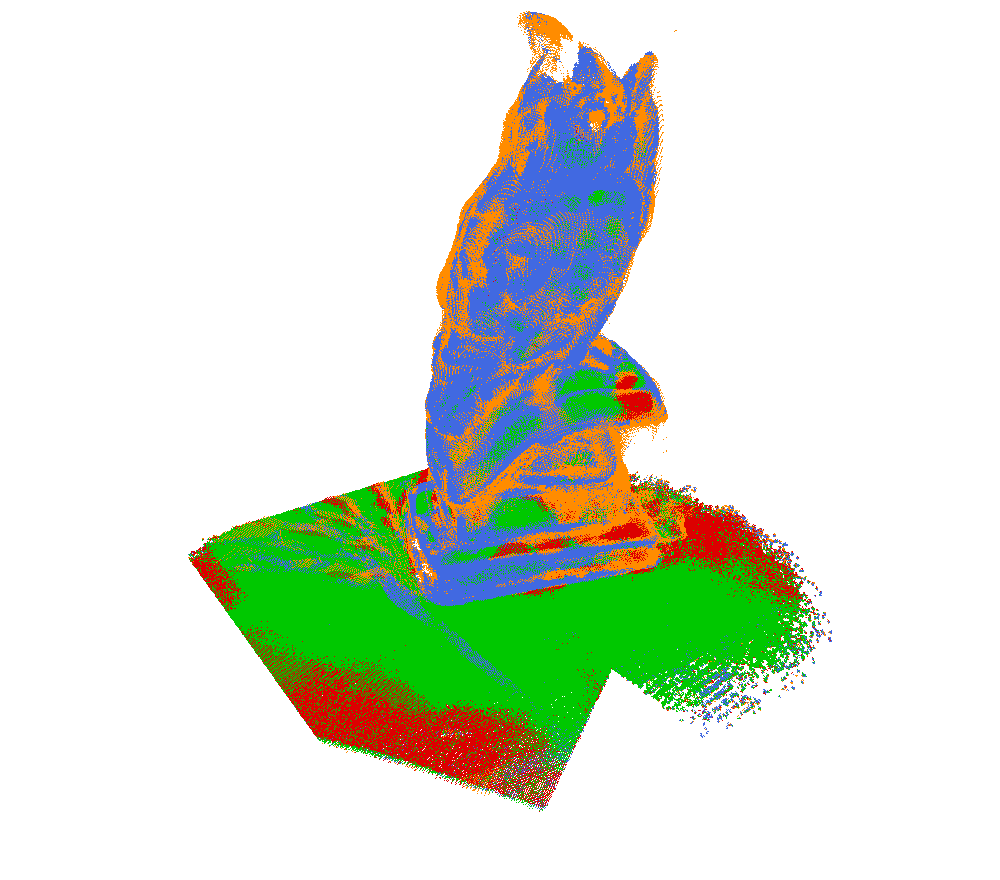}
        \end{subfigure}
    \end{minipage}\\[0.5ex]

    \begin{minipage}[c]{0.1\textwidth}
        \vspace{0.5cm}
        \centering \textbf{VGGT-d, \\ scene 122}
    \end{minipage}%
    \begin{minipage}[c]{0.89\textwidth}
        \begin{subfigure}[t]{0.245\textwidth}
            \includegraphics[width=\textwidth, trim=5cm 8cm 4cm 3cm, clip]{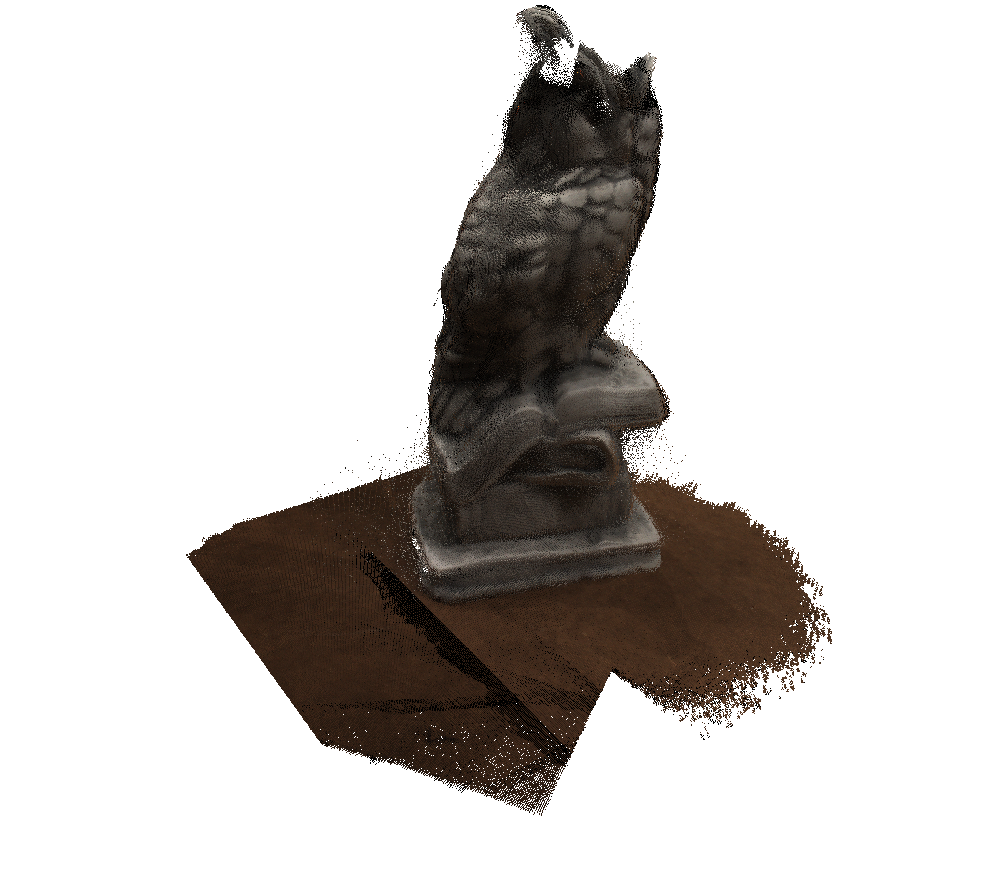}
        \end{subfigure}
        \begin{subfigure}[t]{0.245\textwidth}
            \includegraphics[width=\textwidth, trim=5cm 8cm 4cm 3cm, clip]{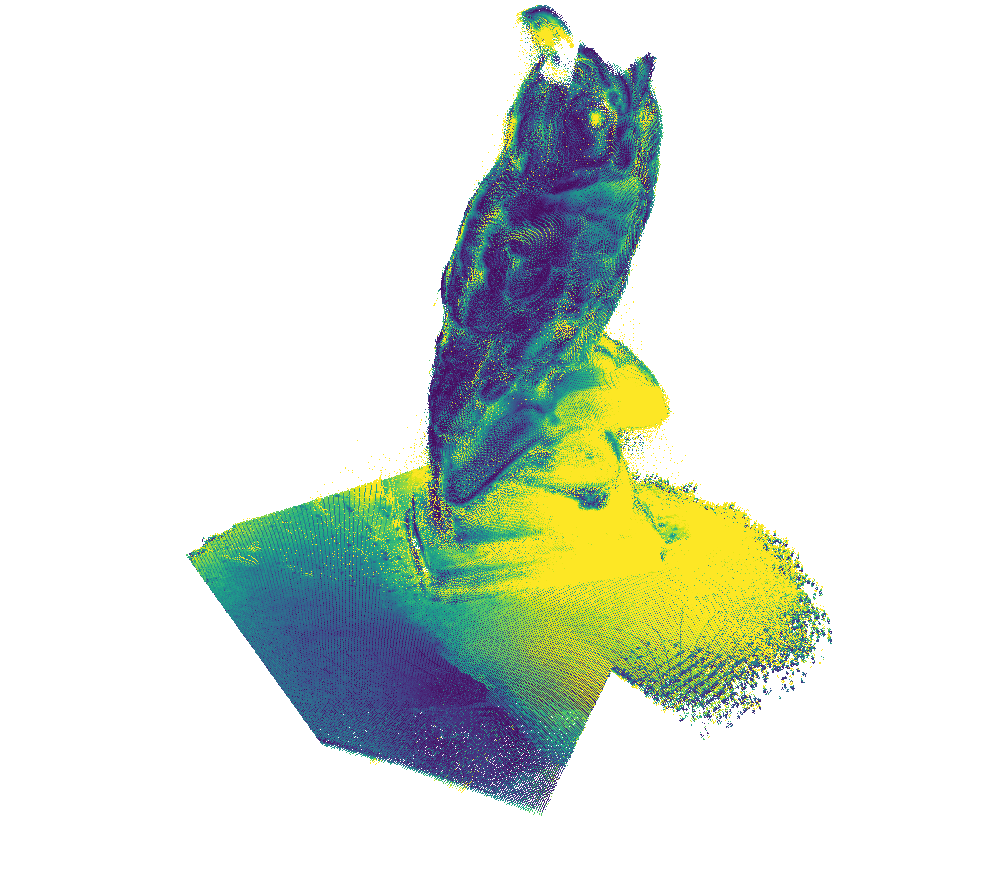}
        \end{subfigure}
        \begin{subfigure}[t]{0.245\textwidth}
            \includegraphics[width=\textwidth, trim=5cm 8cm 4cm 3cm, clip]{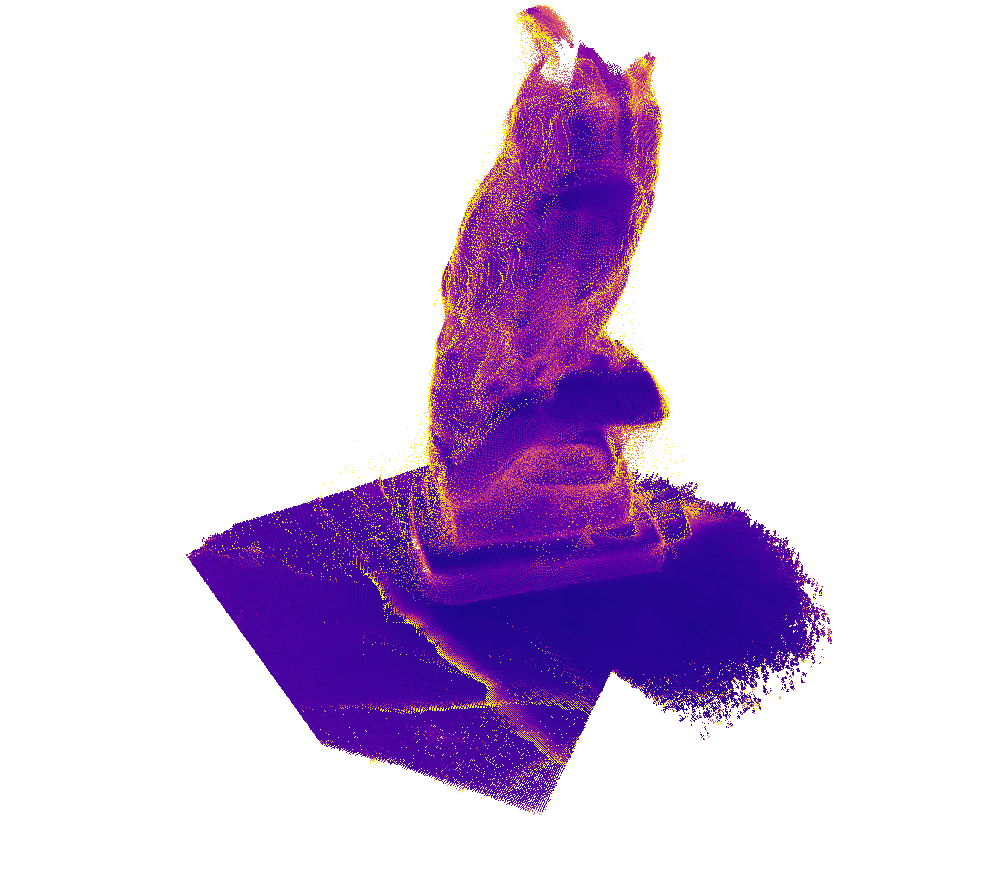}
        \end{subfigure}
        \begin{subfigure}[t]{0.245\textwidth}
            \includegraphics[width=\textwidth, trim=5cm 8cm 4cm 3cm, clip]{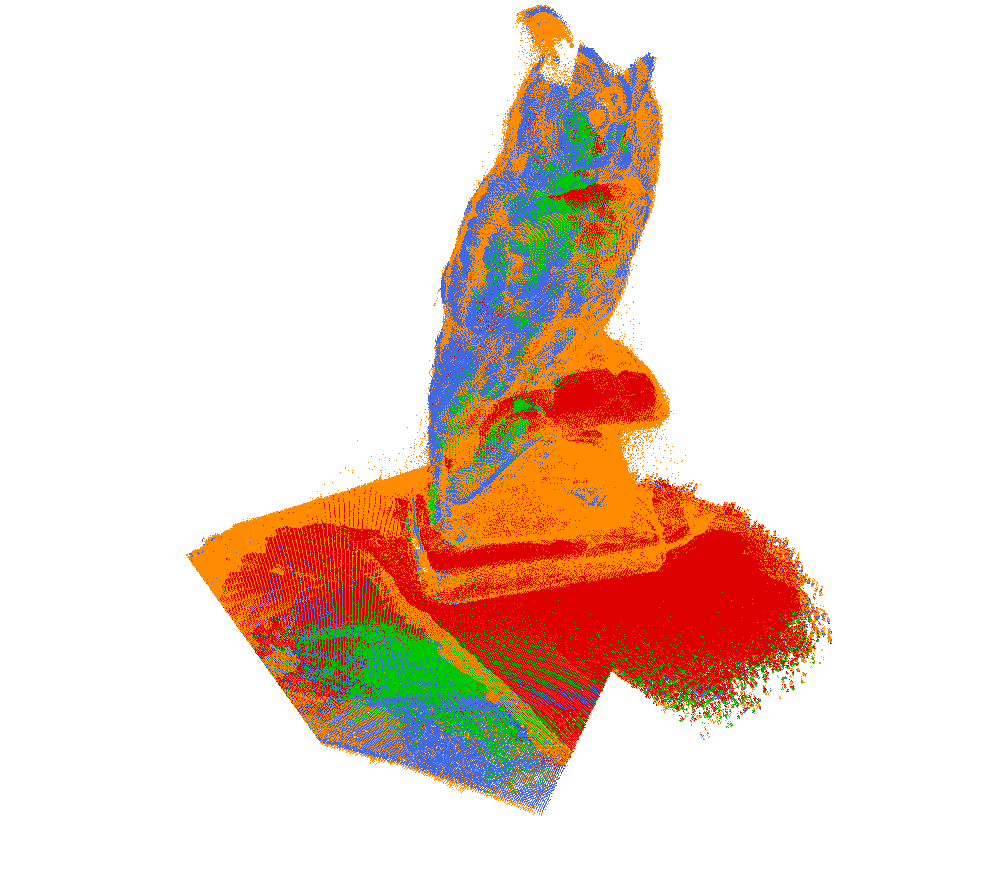}
        \end{subfigure}
    \end{minipage}\\

    \begin{minipage}[c]{0.1\textwidth}
        \vspace{0.5cm}
        \centering \textbf{}
    \end{minipage}%
    \begin{minipage}[c]{0.89\textwidth}
        \makebox[0.245\textwidth]{}
        \begin{subfigure}[t]{0.245\textwidth}
            \includegraphics[width=\textwidth, valign=t]{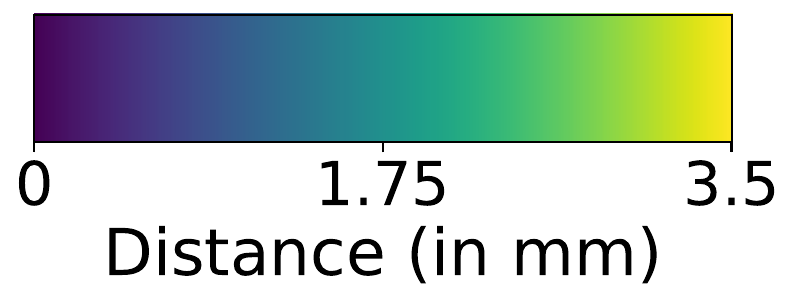}
        \end{subfigure}
        \begin{subfigure}[t]{0.245\textwidth}
            \includegraphics[width=\textwidth, valign=t]{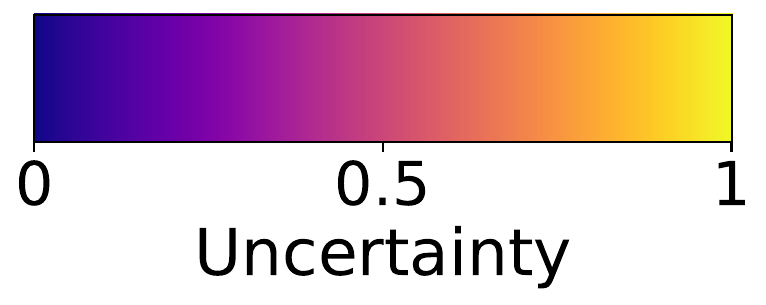}
        \end{subfigure}
        \begin{subfigure}[t]{0.245\textwidth} 
            \centering
            \setlength{\tabcolsep}{2pt}
            \renewcommand{\arraystretch}{1.3} 
            
            \raisebox{-6.5pt}[\height][0pt]{
            \begin{tabular}[t]{%
                >{\centering\arraybackslash}m{0.2\linewidth}
                >{\centering\arraybackslash}m{0.2\linewidth}
                >{\centering\arraybackslash}m{0.2\linewidth}
                >{\centering\arraybackslash}m{0.2\linewidth}
            }
                \cellcolor{ac}\rule{0pt}{1.6em} &
                \cellcolor{ic}\rule{0pt}{1.6em} &
                \cellcolor{au}\rule{0pt}{1.6em} &
                \cellcolor{iu}\rule{0pt}{1.6em} \\
                \vspace{0.05em}
                \small AC & \small IC & \small AU & \small IU \\
            \end{tabular}
            }
        \end{subfigure}
    \end{minipage}

    \caption{Qualitative results of VVGT for scenes 24, 69, and 122 of the DTU dataset.}
    \label{fig:qualitative_DTU_results}
\end{figure*}

Fig.\@~\ref{fig:qualitative_DTU_results} shows the ground truth together with qualitative results of VGGT-p and VGGT-d for scenes 24, 69, and 122. The point clouds are colored based on the predicted color, the pointwise Accuracy (cf.\ Equation~(\ref{eq:acc}) without averaging over the ground truth points), the uncertainty (cf.\ Equation~(\ref{eq:uncertainty})), and the classification of each point into accurate–certain (AC), accurate–uncertain (AU), inaccurate–certain (IC), and inaccurate–uncertain (IU). In this section, we will focus on a discussion of the first two columns of the figure. The uncertainties are discussed in Section \ref{sec:qual_unc}. The predicted point clouds have high visual quality. The black points in scene 24 show minor reconstruction artifacts, whereby the reconstruction with VGGT-p contains more of these artifacts than VGGT-d. These points are artifacts that are caused by the zero padding that is applied by VGGT during cropping. They could therefore be easily filtered out. The coloring with pointwise Accuracy clearly illustrates the higher Accuracy of VGGT-p. Furthermore, Accuracy is good on flat surfaces but tends to be poorer for points directly at edges of the 3D geometry. The Accuracy is slightly worse for the windows in scene 24. Note that these are not real glass windows, as the building shown is a miniature model. 

\subsection{Quality of the Predicted Uncertainties} \label{sec:qual_unc}

Table \ref{tab:vggt_unc_results_per_scene} shows the quality of the predicted uncertainties of VGGT-p and VGGT-d in terms of PAvPU, pAC, pUI, and AUSE. VGGT-p achieves a better uncertainty quality than VGGT-d for all metrics. In particular, the probability that a point is accurate when it is classified as certain by VGGT (pAC) is significantly lower for VGGT-d. VGGT-d thus exhibits the typical phenomenon of overconfidence \citep{guo2017calibration}. However, the quality of the uncertainty prediction of VGGT-p is also suboptimal. For example, the probability that an inaccurate point will be rated as uncertain by the network is only 61.2\%. Therefore, inaccurate points cannot simply be removed based on filtering uncertain points.

\begin{table*}[ht!]
  \centering
  \renewcommand{\arraystretch}{1.06}
  \setlength{\tabcolsep}{4.0pt}
  \caption{Uncertainty quality of VGGT-p (top) and VGGT-d (bottom).}
  \label{tab:vggt_unc_results_per_scene}

  \begin{threeparttable}
      \resizebox{\textwidth}{!}{
          \begin{tabular}{llcccccccccccccccc}
            \toprule
            \textbf{Method} & \textbf{Metric}
              & \textbf{24} & \textbf{37} & \textbf{40} & \textbf{55} & \textbf{63}
              & \textbf{65} & \textbf{69} & \textbf{83} & \textbf{97} & \textbf{105}
              & \textbf{106} & \textbf{110} & \textbf{114} & \textbf{118} & \textbf{122}
              & \textbf{Mean} \\
            \midrule
            \multirow{3}{*}{\textbf{VGGT-p}}
             & PAvPU (in \%) $\uparrow$ & 50.4 & 54.6 & 54.6 & 55.8 & 56.3 & 53.9 & 58.4 & 50.9 & 55.7 & 52.8 & 54.0 & 55.0 & 57.1 & 58.2 & 57.0 & \textbf{55.0} \\
             & pAC (in \%) $\uparrow$& 67.2 & 73.0 & 64.2 & 83.5 & 80.7 & 89.4 & 78.4 & 75.9 & 81.1 & 80.8 & 88.1 & 87.8 & 82.7 & 90.0 & 86.0 & \textbf{80.6} \\
             & pUI (in \%) $\uparrow$& 50.7 & 57.3 & 55.7 & 63.0 & 62.4 & 63.4 & 64.0 & 51.7 & 61.6 & 56.4 & 62.6 & 64.6 & 64.6 & 72.5 & 66.7 & \textbf{61.}2 \\
             & AUSE $\downarrow$ & 0.47 & 0.46 & 0.46 & 0.28 & 0.36 & 0.26 & 0.34 & 0.45 & 0.34 & 0.39 & 0.26 & 0.26 & 0.31 & 0.24 & 0.24 & \textbf{0.34} \\
            \addlinespace[2pt] 
            \hdashline[1pt/2pt]
            \addlinespace[2pt]
            \multirow{3}{*}{\textbf{VGGT-d}}
             & PAvPU (in \%) $\uparrow$ & 52.7 & 53.2 & 52.3 & 54.3 & 51.8 & 52.8 & 56.4 & 52.6 & 54.5 & 53.2 & 54.4 & 59.7 & 60.0 & 53.7 & 50.3 & 54.1 \\
             & pAC (in \%) $\uparrow$ & 52.4 & 49.9 & 36.9 & 47.2 & 35.3 & 44.5 & 53.5 & 46.9 & 44.6 & 50.4 & 53.9 & 56.8 & 66.4 & 49.9 & 38.9 & 48.5 \\
             & pUI (in \%) $\uparrow$ & 52.6 & 53.0 & 51.8 & 53.7 & 51.3 & 52.4 & 56.0 & 52.4 & 53.7 & 53.0 & 54.4 & 59.2 & 61.4 & 53.5 & 50.2 & 53.9 \\
             & AUSE $\downarrow$ & 0.69 & 0.76 & 0.79 & 0.74 & 0.90 & 0.79 & 0.60 & 0.86 & 0.77 & 0.72 & 0.62 & 0.55 & 0.44 & 0.62 & 0.89 & 0.72 \\
            \bottomrule
          \end{tabular}
        }
  \end{threeparttable}
\end{table*}

As mentioned, Fig.\@~\ref{fig:qualitative_DTU_results} also visualizes the uncertainty predictions and quality. As a reminder, the most important aspects for meaningful uncertainties are that (1) points with low uncertainty should be accurate and (2) inaccurate points should be uncertain. The right column of Fig.\@~\ref{fig:qualitative_DTU_results} can be interpreted as follows: Many green and orange points, but few red points indicate high uncertainty quality. Many green and blue points, but few orange points indicate good accuracy. Red points (inaccurate but certain) are the most problematic, as overconfidence can lead to critical misjudgments. The figure shows that the estimated uncertainties are suboptimal. For example, VGGT-d often provides a low uncertainty for points on the ground, even though these are often inaccurate. But even with VGGT-p, there are inaccurate points that are considered certain, e.g., in scene 24 on the roof on the left or in scene 69 on the hat of the left snowman.
In general, VGGT-d tends to report lower uncertainty estimates compared to VGGT-p, despite being less accurate. This underscores the above statement about the overconfidence of VGGT-d. We suspect that the predominant source of this overconfidence arises from VGGT providing uncertainty estimates only for depth maps, but not for the interior and exterior orientation. However, interior and exterior orientation are required to compute the point cloud from the depth maps. 
Interestingly, VGGT is always very uncertain about the windows in scene 24, even though the accuracy there is only slightly below average. We suspect that the reason for this lies in contextual knowledge of VGGT: During training, the model learned that windows are generally difficult to predict. However, the building model of scene 24 does not contain real glass windows, and the accuracy is actually better than the uncertainty would suggest.

\begin{figure}[ht!]
    \centering
    \includegraphics[width=0.95\linewidth]{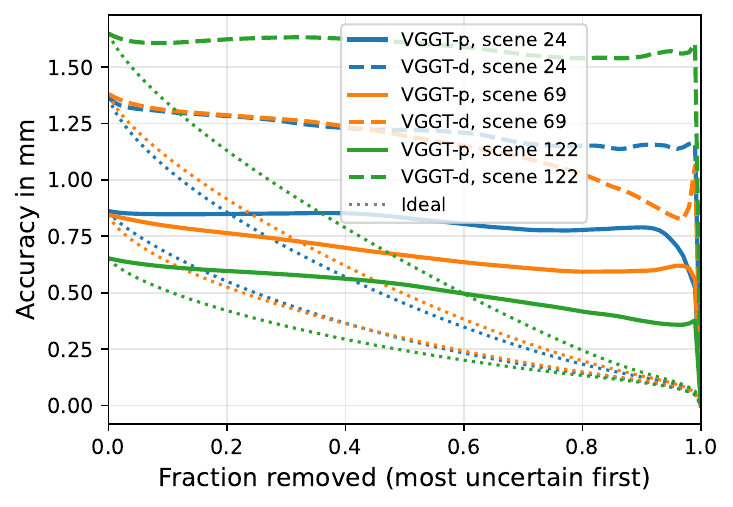}
    \caption{Sparsification curves of VGGT-p and VGGT-d for scenes 24, 69, and 122. 
    }
    \label{fig:sparsification_curves}
\end{figure}

Fig.\@~\ref{fig:sparsification_curves} shows the sparsification curves for VGGT-p and VGGT-d for three scenes. When points are filtered out based on uncertainty, accuracy improves only slowly for both VGGT-p and VGGT-d. This clearly illustrates that the correlation between uncertainty and accuracy is suboptimal. Consequently, there is room for improvement in terms of uncertainty quality, which should be addressed in future research. 

\begin{figure}[ht!]
    \centering
    \includegraphics[width=0.9\linewidth]{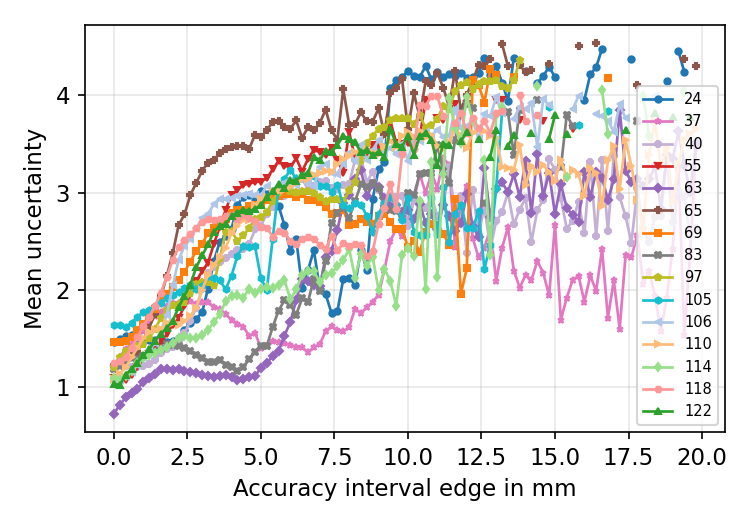}
    \caption{Correlation Curves of VGGT-d for all DTU scenes. The lines are interrupted when there are no points in an Accuracy interval.}
    \label{fig:inverted_sparsification_curves}
\end{figure}

Finally, Fig.\ \ref{fig:inverted_sparsification_curves} visualizes the correlation curves. Ideally, the curves should rise linearly (with an arbitrary slope, since the uncertainties are unscaled). This is the case for points with a good Accuracy ($<$2\,mm). However, the linear relationship does not apply to points with poorer Accuracy for many scenes. There are not many points with very poor Accuracy ($>$10\,mm), which is why the curves in this area are very noisy.

\section{Conclusion}\label{sec:Conclusion}

We analyze the uncertainty quality of the depth maps and point maps predicted by VGGT on the DTU benchmark dataset. Our evaluations show that a confidence threshold of 2.0 is a good choice for filtering the raw output of VGGT. Since we have both a causal justification and empirical evidence from the evaluation on the DTU dataset, we believe that this threshold can be generally recommended as a starting point for potential dataset-specific fine tuning, since it does not necessarily lead to the best results for every single scene. Depending on specific application requirements, a different trade-off between accuracy and completeness may be more appropriate, which also affects the choice of the confidence threshold (higher for a focus on accuracy, lower for a focus on completeness). Moreover, we find that the point map branch is more accurate than the depth map branch for 3D reconstruction, when the feed-forward results are evaluated (i.e., no bundle adjustment in post-processing). The quality of the predicted uncertainties is also significantly better for the point map branch, while the depth map branch suffers from overconfidence in particular.

The analysis shows that improving the uncertainty prediction of VGGT has high potential to improve the accuracy of the 3D reconstruction. In principle, the output of VGGT is redundant, as it outputs a 3D point for each pixel. If the uncertainties were of higher quality, they would have more potential to increase the 3D reconstruction accuracy, e.g., by filtering uncertain (and therefore inaccurate) points, or by utilizing them for multi-view fusion. Potentially, higher uncertainty quality could be achieved by taking into account the epistemic uncertainty component. Finally, metric uncertainty quantification per predicted 3D point is currently not yet possible, but would be advantageous for many photogrammetric and metrology tasks and safety-critical applications.

{
	\begin{spacing}{1.17}
		\normalsize
		\bibliography{main} 

@inproceedings{schonberger2016structure,
  title={Structure-from-motion revisited},
  author={Schönberger, Johannes L and Frahm, Jan-Michael},
  booktitle={Proceedings of the IEEE Conference on Computer Vision and Pattern Recognition},
  pages={4104--4113},
  year={2016}
}

@article{aanaes2016large,
  title={Large-scale data for multiple-view stereopsis},
  author={Aan{\ae}s, Henrik and Jensen, Rasmus Ramsb{\o}l and Vogiatzis, George and Tola, Engin and Dahl, Anders Bjorholm},
  journal={International Journal of Computer Vision},
  volume={120},
  pages={153--168},
  year={2016},
  publisher={Springer}
}

@inproceedings{detone2018superpoint,
  title={Superpoint: Self-supervised interest point detection and description},
  author={DeTone, Daniel and Malisiewicz, Tomasz and Rabinovich, Andrew},
  booktitle={Proceedings of the IEEE conference on computer vision and pattern recognition workshops},
  pages={224--236},
  year={2018}
}

@inproceedings{sarlin2020superglue,
  title={Superglue: Learning feature matching with graph neural networks},
  author={Sarlin, Paul-Edouard and DeTone, Daniel and Malisiewicz, Tomasz and Rabinovich, Andrew},
  booktitle={Proceedings of the IEEE/CVF conference on computer vision and pattern recognition},
  pages={4938--4947},
  year={2020}
}

@article{zhang2025review,
  title={Review of Feed-forward 3D Reconstruction: From DUSt3R to VGGT},
  author={Zhang, Wei and Wu, Yihang and Li, Songhua and Ma, Wenjie and Ma, Xin and Li, Qiang and Wang, Qi},
  journal={arXiv preprint arXiv:2507.08448},
  year={2025}
}

@inproceedings{wang2024dust3r,
  title={Dust3r: Geometric 3d vision made easy},
  author={Wang, Shuzhe and Leroy, Vincent and Cabon, Yohann and Chidlovskii, Boris and Revaud, Jerome},
  booktitle={Proceedings of the IEEE/CVF Conference on Computer Vision and Pattern Recognition},
  pages={20697--20709},
  year={2024}
}

@inproceedings{leroy2024grounding,
  title={Grounding image matching in 3d with mast3r},
  author={Leroy, Vincent and Cabon, Yohann and Revaud, J{\'e}r{\^o}me},
  booktitle={European Conference on Computer Vision},
  pages={71--91},
  year={2024},
  organization={Springer}
}

@inproceedings{wang2025vggt,
  title={Vggt: Visual geometry grounded transformer},
  author={Wang, Jianyuan and Chen, Minghao and Karaev, Nikita and Vedaldi, Andrea and Rupprecht, Christian and Novotny, David},
  booktitle={Proceedings of the Computer Vision and Pattern Recognition Conference},
  pages={5294--5306},
  year={2025}
}

@article{vaswani2017attention,
  title={Attention is all you need},
  author={Vaswani, Ashish and Shazeer, Noam and Parmar, Niki and Uszkoreit, Jakob and Jones, Llion and Gomez, Aidan N and Kaiser, {\L}ukasz and Polosukhin, Illia},
  journal={Advances in neural information processing systems},
  volume={30},
  year={2017}
}

@article{zhang2025advances,
  title={Advances in feed-forward 3d reconstruction and view synthesis: A survey},
  author={Zhang, Jiahui and Li, Yuelei and Chen, Anpei and Xu, Muyu and Liu, Kunhao and Wang, Jianyuan and Long, Xiao-Xiao and Liang, Hanxue and Xu, Zexiang and Su, Hao and others},
  journal={arXiv preprint arXiv:2507.14501},
  year={2025}
}

@inproceedings{tang2025mv,
  title={Mv-dust3r+: Single-stage scene reconstruction from sparse views in 2 seconds},
  author={Tang, Zhenggang and Fan, Yuchen and Wang, Dilin and Xu, Hongyu and Ranjan, Rakesh and Schwing, Alexander and Yan, Zhicheng},
  booktitle={Proceedings of the Computer Vision and Pattern Recognition Conference},
  pages={5283--5293},
  year={2025}
}

@inproceedings{yang2025fast3r,
  title={Fast3r: Towards 3d reconstruction of 1000+ images in one forward pass},
  author={Yang, Jianing and Sax, Alexander and Liang, Kevin J and Henaff, Mikael and Tang, Hao and Cao, Ang and Chai, Joyce and Meier, Franziska and Feiszli, Matt},
  booktitle={Proceedings of the Computer Vision and Pattern Recognition Conference},
  pages={21924--21935},
  year={2025}
}

@article{kendall2017uncertainties,
  title={What uncertainties do we need in bayesian deep learning for computer vision?},
  author={Kendall, Alex and Gal, Yarin},
  journal={Advances in neural information processing systems},
  volume={30},
  year={2017}
}

@article{gawlikowski2023survey,
  title={A survey of uncertainty in deep neural networks},
  author={Gawlikowski, Jakob and Tassi, Cedrique Rovile Njieutcheu and Ali, Mohsin and Lee, Jongseok and Humt, Matthias and Feng, Jianxiang and Kruspe, Anna and Triebel, Rudolph and Jung, Peter and Roscher, Ribana and others},
  journal={Artificial Intelligence Review},
  volume={56},
  number={Suppl 1},
  pages={1513--1589},
  year={2023},
  publisher={Springer}
}

@article{der2009aleatory,
  title={Aleatory or epistemic? Does it matter?},
  author={Der Kiureghian, Armen and Ditlevsen, Ove},
  journal={Structural safety},
  volume={31},
  number={2},
  pages={105--112},
  year={2009},
  publisher={Elsevier}
}

@article{he2023survey,
  title={A survey on uncertainty quantification methods for deep learning},
  author={He, Wenchong and Jiang, Zhe and Xiao, Tingsong and Xu, Zelin and Li, Yukun},
  journal={arXiv preprint arXiv:2302.13425},
  year={2023}
}

@inproceedings{gal2016dropout,
  title={Dropout as a bayesian approximation: Representing model uncertainty in deep learning},
  author={Gal, Yarin and Ghahramani, Zoubin},
  booktitle={International Conference on Machine Learning},
  pages={1050--1059},
  year={2016},
  organization={PMLR}
}

@book{neal2012bayesian,
  title={{Bayesian} {Learning} for {Neural} {Networks}},
  author={Neal, Radford M},
  volume={118},
  year={2012},
  publisher={Springer Science \& Business Media}
}

@article{lakshminarayanan2017simple,
  title={Simple and scalable predictive uncertainty estimation using deep ensembles},
  author={Lakshminarayanan, Balaji and Pritzel, Alexander and Blundell, Charles},
  journal={Advances in neural information processing systems},
  volume={30},
  year={2017}
}

@article{ganaie2022ensemble,
  title={Ensemble deep learning: A review},
  author={Ganaie, Mudasir A and Hu, Minghui and Malik, Ashwani Kumar and Tanveer, Muhammad and Suganthan, Ponnuthurai N},
  journal={Engineering Applications of Artificial Intelligence},
  volume={115},
  pages={105151},
  year={2022},
  publisher={Elsevier}
}

@article{zhou2024comprehensive,
  title={A comprehensive review of vision-based 3d reconstruction methods},
  author={Zhou, Linglong and Wu, Guoxin and Zuo, Yunbo and Chen, Xuanyu and Hu, Hongle},
  journal={Sensors},
  volume={24},
  number={7},
  pages={2314},
  year={2024},
  publisher={MDPI}
}

@book{hartley2003multiple,
  title={Multiple view geometry in computer vision},
  author={Hartley, Richard and Zisserman, Andrew},
  year={2003},
  publisher={Cambridge university press}
}

@inproceedings{seitz2006comparison,
  title={A comparison and evaluation of multi-view stereo reconstruction algorithms},
  author={Seitz, Steven M and Curless, Brian and Diebel, James and Scharstein, Daniel and Szeliski, Richard},
  booktitle={2006 IEEE computer society conference on computer vision and pattern recognition (CVPR'06)},
  volume={1},
  pages={519--528},
  year={2006},
  organization={IEEE}
}

@article{stathopoulou2023survey,
  title={A survey on conventional and learning-based methods for multi-view stereo},
  author={Stathopoulou, Elisavet Konstantina and Remondino, Fabio},
  journal={The Photogrammetric Record},
  volume={38},
  number={183},
  pages={374--407},
  year={2023},
  publisher={Wiley Online Library}
}

@article{liu2025survey,
  title={A Survey of 3D Reconstruction: The Evolution from Multi-View Geometry to NeRF and 3DGS},
  author={Liu, Shuai and Yang, Mengmeng and Xing, Tingyan and Yang, Ran},
  journal={Sensors},
  volume={25},
  number={18},
  pages={5748},
  year={2025},
  publisher={MDPI}
}

@article{wu2025evaluation,
  title={An Evaluation of DUSt3R/MASt3R/VGGT 3D Reconstruction on Photogrammetric Aerial Blocks},
  author={Wu, Xinyi and Landgraf, Steven and Ulrich, Markus and Qin, Rongjun},
  journal={arXiv preprint arXiv:2507.14798},
  year={2025}
}

@inproceedings{schonberger2016pixelwise,
  title={Pixelwise view selection for unstructured multi-view stereo},
  author={Sch{\"o}nberger, Johannes L and Zheng, Enliang and Frahm, Jan-Michael and Pollefeys, Marc},
  booktitle={European conference on computer vision},
  pages={501--518},
  year={2016},
  organization={Springer}
}

@inproceedings{li2023neuralangelo,
  title={Neuralangelo: High-fidelity neural surface reconstruction},
  author={Li, Zhaoshuo and M{\"u}ller, Thomas and Evans, Alex and Taylor, Russell H and Unberath, Mathias and Liu, Ming-Yu and Lin, Chen-Hsuan},
  booktitle={Proceedings of the IEEE/CVF Conference on Computer Vision and Pattern Recognition},
  pages={8456--8465},
  year={2023}
}

@article{chen2024pgsr,
  title={PGSR: Planar-based Gaussian Splatting for Efficient and High-fidelity Surface Reconstruction},
  author={Chen, Danpeng and Li, Hai and Ye, Weicai and Wang, Yifan and Xie, Weijian and Zhai, Shangjin and Wang, Nan and Liu, Haomin and Bao, Hujun and Zhang, Guofeng},
  journal={IEEE Transactions on Visualization and Computer Graphics},
  year={2024},
  publisher={IEEE}
}

@article{mukhoti2018evaluating,
  title={Evaluating bayesian deep learning methods for semantic segmentation},
  author={Mukhoti, Jishnu and Gal, Yarin},
  journal={arXiv preprint arXiv:1811.12709},
  year={2018}
}

@inproceedings{ilg2018uncertainty,
  title={Uncertainty estimates and multi-hypotheses networks for optical flow},
  author={Ilg, Eddy and Cicek, Ozgun and Galesso, Silvio and Klein, Aaron and Makansi, Osama and Hutter, Frank and Brox, Thomas},
  booktitle={Proceedings of the European Conference on Computer Vision (ECCV)},
  pages={652--667},
  year={2018}
}

@inproceedings{huang20242dgs,
  title={2d Gaussian Splatting for Geometrically Accurate Radiance Fields},
  author={Huang, Binbin and Yu, Zehao and Chen, Anpei and Geiger, Andreas and Gao, Shenghua},
  booktitle={ACM SIGGRAPH 2024},
  pages={1--11},
  year={2024}
}

@article{yariv2020multiview,
  title={Multiview neural surface reconstruction by disentangling geometry and appearance},
  author={Yariv, Lior and Kasten, Yoni and Moran, Dror and Galun, Meirav and Atzmon, Matan and Ronen, Basri and Lipman, Yaron},
  journal={Advances in Neural Information Processing Systems},
  volume={33},
  pages={2492--2502},
  year={2020}
}

@inproceedings{wang2024vggsfm,
  title={VGGSfM: Visual geometry grounded deep structure from motion},
  author={Wang, Jianyuan and Karaev, Nikita and Rupprecht, Christian and Novotny, David},
  booktitle={Proceedings of the IEEE/CVF conference on computer vision and pattern recognition},
  pages={21686--21697},
  year={2024}
}

@article{landgraf2025comparative,
  title={A comparative study on multi-task uncertainty quantification in semantic segmentation and monocular depth estimation},
  author={Landgraf, Steven and Hillemann, Markus and Kapler, Theodor and Ulrich, Markus},
  journal={tm-Technisches Messen},
  volume={92},
  number={7-8},
  pages={298--310},
  year={2025},
  publisher={Oldenbourg Wissenschaftsverlag}
}

@article{oquab2024dinov2,
  title={DINOv2: Learning Robust Visual Features without Supervision},
  author={Oquab, Maxime and Darcet, Timoth{\'e}e and Moutakanni, Th{\'e}o and Vo, Huy and Szafraniec, Marc and Khalidov, Vasil and Fernandez, Pierre and Haziza, Daniel and Massa, Francisco and El-Nouby, Alaaeldin and others},
  journal={Transactions on Machine Learning Research Journal},
  pages={1--31},
  year={2024}
}

@inproceedings{guo2017calibration,
  title={On calibration of modern neural networks},
  author={Guo, Chuan and Pleiss, Geoff and Sun, Yu and Weinberger, Kilian Q},
  booktitle={International conference on machine learning},
  pages={1321--1330},
  year={2017},
  organization={PMLR}
}

@article{landgraf2025rethinking,
  title={Rethinking Semi-supervised Segmentation Beyond Accuracy: Reliability and Robustness},
  author={Landgraf, Steven and Hillemann, Markus and Ulrich, Markus},
  journal={arXiv preprint arXiv:2506.05917},
  year={2025}
}
	\end{spacing}
}

\end{document}